\documentclass[11pt]{article}

\usepackage[final]{acl}

\usepackage{times}
\usepackage{latexsym}

\usepackage[T1]{fontenc}

\usepackage[utf8]{inputenc}

\usepackage{microtype}

\usepackage{inconsolata}

\usepackage{graphicx}

\usepackage{float}
\usepackage{fancyvrb}
\usepackage{graphicx}
\usepackage{boldline} 
\usepackage{booktabs}
\usepackage{multirow}
\usepackage{sidecap}
\usepackage{subcaption}
\usepackage{amsmath}
\usepackage{amsfonts}
\usepackage{xcolor}
\usepackage{bm}
\usepackage{hyperref}
\usepackage{rotating}
\usepackage{adjustbox}
\usepackage{algorithm}
\usepackage{algpseudocode} 
\usepackage{array}
\definecolor{green}{rgb}{0.0, 0.5, 0.0}
\definecolor{red}{rgb}{0.82, 0.1, 0.26}
\definecolor{codegray}{gray}{0.9}

\usepackage{pifont}

\newlength{\Oldarrayrulewidth}

\definecolor{darkpurple}{RGB}{75, 0, 130}
\definecolor{darkyellow}{RGB}{230, 184, 0}

\usepackage[most]{tcolorbox}
\tcbuselibrary{breakable}   

\newtcolorbox{myshadowbox}{
    enhanced,
    colback=white, 
    colframe=black, 
    shadow={1mm}{-1mm}{0mm}{black!50!white}, 
    boxrule=0.5pt 
}
\usepackage{colortbl} 
\usepackage{tabularx} 

\newcommand{\rdn}{\textcolor{red}{$\downarrow$}}
\newcommand{\gup}{\textcolor{green}{$\uparrow$}}
\newcounter{prop}
\newcommand{\proposition}[1]{%
  \refstepcounter{prop}%
  \noindent \textbf{Proposition~\theprop.} \label{#1}%
}

\newif\ifcoloredtext
\coloredtextfalse 
\newcommand{\red}[1]{%
  \ifcoloredtext
    \textcolor{red}{#1}%
  \else
    #1%
  \fi
}

\usepackage{amsthm}
\newtheorem*{assumption*}{Assumption}

\title{Chain-in-Tree: Back to Sequential Reasoning in LLM Tree Search}

\author{Xinzhe Li \\
  Independent Researcher \\
\texttt{sergioli212@outlook.com}\\}

\begin{document}
\maketitle

\begin{abstract}
Test-time scaling improves large language models (LLMs) on long-horizon reasoning tasks by allocating more compute at inference.
LLM inference via tree search (LITS) achieves strong performance but is highly inefficient.
We propose \textbf{Chain-in-Tree (CiT)}, a plug-in framework that decides \emph{when} to branch during search instead of expanding at every step.
CiT introduces lightweight \textbf{Branching Necessity (BN)} evaluations, including \textbf{BN-DP} (direct prompting) and \textbf{BN-SC} (self-consistency).
Integrated into Tree of Thoughts, ReST-MCTS, and RAP, BN-DP reduces token generation, model calls, and runtime by \textbf{75--85\%} on GSM8K and Math500, with often negligible or no accuracy loss.
BN-SC typically yields substantial savings (up to 80\%) generally but shows instability in 1--4 out of 14 settings, caused by a small subset of examples that produce extremely long reasoning steps.
We theoretically prove that BN-DP never increases policy invocations and release unified implementations applicable across LITS frameworks.
The full codebase is publicly available at \url{https://github.com/xinzhel/chain_in_tree}.
\end{abstract}





\section{Introduction}
Large language models (LLMs) excel at tasks such as mathematical and commonsense reasoning, but their performance often improves further when additional test-time compute is allocated.  
Recent work has shown that \emph{test-time scaling} enables LLMs to explore multiple reasoning paths, thereby making better use of knowledge they already possess \citep{sprague2025to}.  
Among test-time scaling methods, tree-search–based approaches have achieved state-of-the-art results on benchmarks like GSM8K and Math500, by treating reasoning as sequential decision-making and exploring multiple candidate trajectories \citep{zhang2024restmcts,yao2023tree}.  
Such approaches pivot LLMs—otherwise weak in zero-shot planning—toward effective performance on planning problems \citep{valmeekam2023planbench}.  

However, tree-search–based frameworks are highly inefficient: compared to simpler iterative prompting methods, they are often 10--20 times slower \citep{chen-etal-2024-tree}.  
A central limitation is that they operate at a fixed granularity of reasoning steps.  
In some cases, this granularity is enforced explicitly, for example by treating each sub-question–answer pair as a node \citep{hao-etal-2023-reasoning}.  
Other approaches, such as MCTS with free-form thoughts \citep{zhang2024restmcts,yao2023tree}, allow more flexibility but still assume that every generated thought must branch independently.  
In practice, however, a reasoning step in mathematics may correspond to a single operation or a tightly coupled set of operations, while in domains with deterministic action spaces (e.g., board games) the step granularity is even more strictly defined.  
This rigidity forces branching even on trivial steps, resulting in excessive LLM calls during both expansion and simulation.  
We address this limitation by proposing \textbf{Chain-in-Tree (CiT)}, a plug-in for LLM-in-the-loop tree search that adaptively decides \emph{when} branching is necessary.  

\begin{figure*}[ht!]
    \centering

    \begin{subfigure}{0.48\textwidth}
        \centering
        \includegraphics[width=\textwidth]{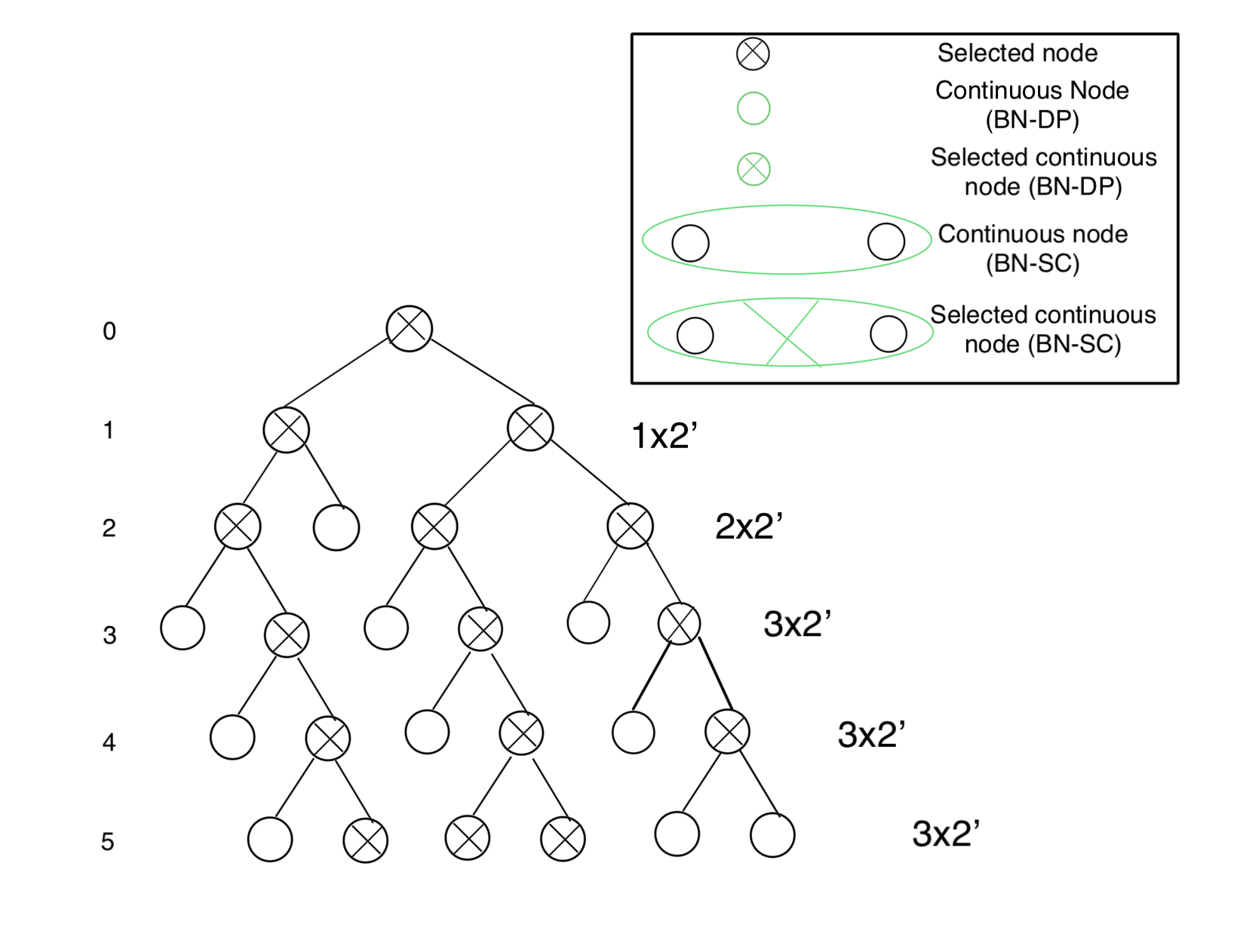}
        \caption{ToT-Beam Search.}
        \label{fig:bfs}
    \end{subfigure}
    \hfill
    \begin{subfigure}{0.48\textwidth}
        \centering
        \includegraphics[width=\textwidth]{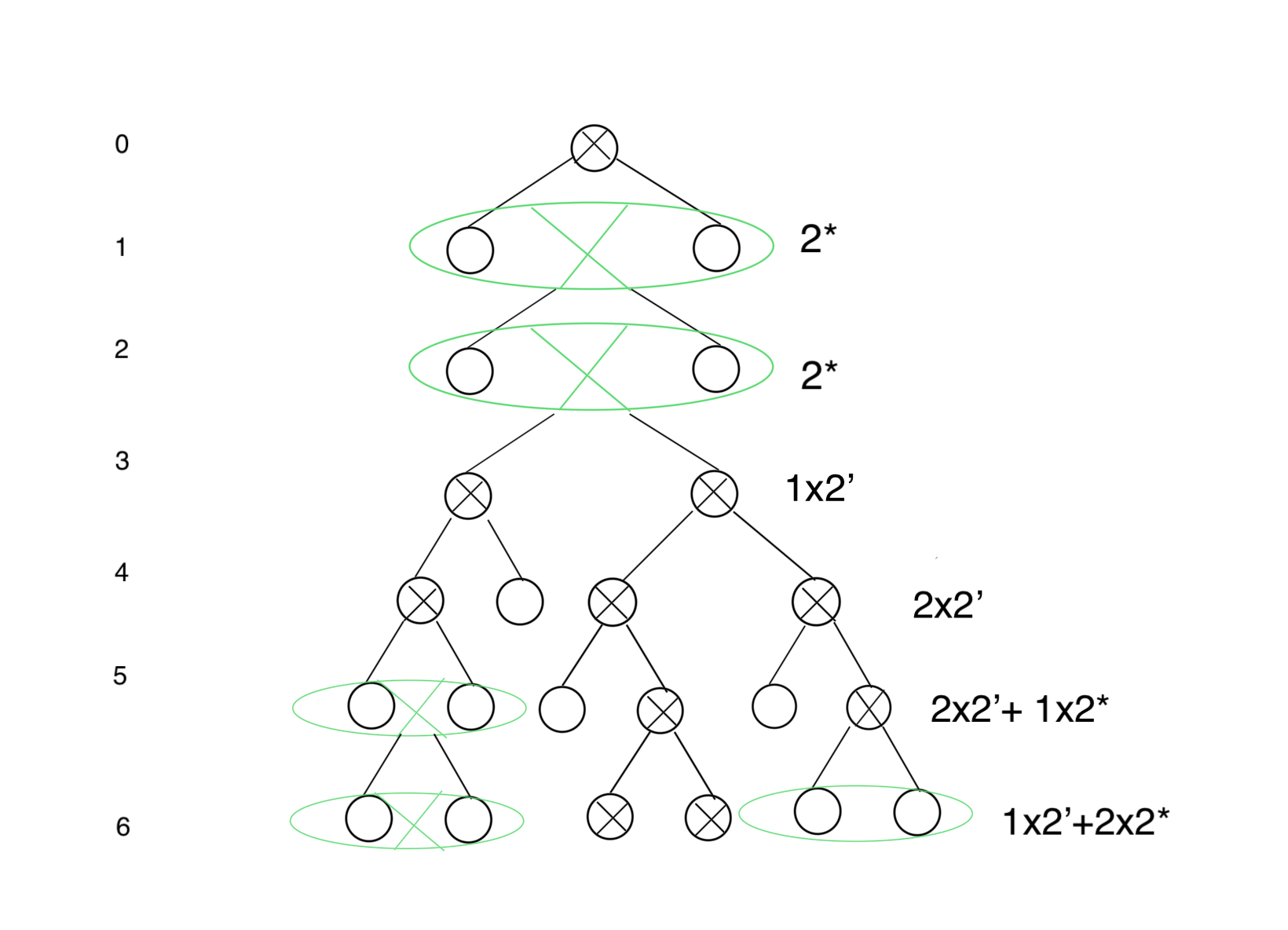}
        \caption{ToT-Beam Search + Chaining (BN-SC).}
        \label{fig:bfs_bndp}
    \end{subfigure}
    \caption{Tree-of-Thoughts Beam Search (ToT-BS) vs. ToT-BS with Chaining (BN-SC). Primes $\prime$ indicate beam sampling size; asterisks $*$ indicate BN-judge sampling size. See Appendix \ref{app:theoretical_costs} for BN-DP.}
    \label{fig:bfs_bn}
\end{figure*}

The key idea is to introduce continuous nodes chained together: steps deemed confident or routine by the model are chained sequentially, while branching is reserved only for uncertain points where exploration is valuable. 
\red{Figure \ref{fig:bfs_bn} provides a high-level illustration of how CiT differs from an existing LITS framework.}
This reduces unnecessary expansion, lowers inference costs, and preserves the search capacity of existing frameworks.  

\paragraph{Contributions.}  
Our contributions are as follows:

\begin{itemize}
    \item We formulate the decision of when to branch as a new component in tree search and propose two lightweight \textbf{Branching Necessity (BN) evaluation methods}: Direct Prompting (BN-DP) and Self-Consistency (BN-SC). These methods allow LLMs to autonomously determine whether branching is needed at each step. \red{The algorithmic design enabled by introducing this additional phase is detailed in Appendix~\ref{app:novelty_clarification}.}

    \item We provide a \textbf{theoretical guarantee} that BN-DP never increases runtime relative to the baseline, thereby ensuring efficiency by design.  
    \item We conduct a comprehensive empirical study across 14 settings: two mathematical reasoning benchmarks (GSM8K, Math500), two base LLMs (Qwen3-32B, LLaMA3-8B) serving either as search policies or BN evaluators, and three representative LITS frameworks (ToT-BS, ReST-MCTS, RAP).  

    \item Our experiments show that \textbf{BN-DP consistently reduces runtime by 75--85\% across all settings with negligible accuracy loss and in some cases improved accuracy}, confirming its reliability.  
    In contrast, \textbf{BN-SC achieves substantial savings in most settings but exhibits instability in a few cases}.  
    Finally, we demonstrate that the overall effectiveness of CiT strongly depends on the accuracy of BN evaluation.

    \item We release a \textbf{unified implementation} of ToT-BS, RAP, and ReST-MCTS with modular LLM-profiled roles, enabling consistent comparison across frameworks and simplifying extension of CiT to future LITS variants.  
\end{itemize}

\section{Related Work}
We situate our method within a unified view of LLM inference via tree search (LITS), showing how our plug-and-play module can be applied across different frameworks.  
In addition, we discuss two related lines of work that share a similar high-level motivation—reducing unnecessary branching—but operate in different settings and with distinct mechanisms.  



\paragraph{LLM Inference via Tree Search (LITS).}
A growing line of work has explored performing LLM inference through tree-search (LITS) \citep{hao-etal-2023-reasoning, yao2023tree}.  
Following the unified view of \citet{li2025a}, these frameworks can be cast into a Markov Decision Process (MDP), where backbone LLMs are profiled into three main roles:  
(1) \textbf{Policy}: generative LLMs produce candidate actions $a_t$ (e.g., reasoning steps) given a state $s_t$ (e.g., the task with prior steps). This usage traces back to the ReAct paradigm \citep{yao2023react};  
(2) \textbf{Reward model}: either using generative LLMs to score candidate steps \citep{yao2023tree, hao-etal-2023-reasoning} or employing dedicated non-generative/fine-tuned models \citep{dai2025process, luo2025improve};  
(3) \textbf{Transition model}: explicitly predicting the next state $s_{t+1}$ and its confidence $r_\text{conf}$ when $a_t$ is executed (e.g., RAP \citep{hao-etal-2023-reasoning}), or more commonly, updating the state by concatenating new thoughts (e.g., ToT \citep{yao2023tree}, ReST-MCTS \citep{zhang2024restmcts}), in which case $r_\text{conf}=1$.  
Our experiments span both types of reward models and both explicit and concatenation-based transition models, demonstrating that CiT is broadly compatible across frameworks.  

\paragraph{Tree Search Efficiency}
\red{Broadly, prior work on improving tree-search efficiency for LLMs can be grouped into two orthogonal and complementary directions.}
The first direction, to which \textbf{CiT} belongs, performs \emph{search-level control} by reducing unnecessary branching or reasoning steps.
\red{One concurrent line of work} is \citet{li2025entropy}, who proposed branching-on-demand in beam search decoding based on entropy over token-level probability distributions.  
While the high-level motivation is similar—branching only when necessary—the setting is fundamentally different.  
Their method operates at the token level in traditional sequence generation tasks such as machine translation and speech recognition \citep{sutskever2014sequence}, where beam search has long been standard.  
\red{FETCH \citep{wang-etal-2025-dont} tackles a related but different decision problem. 
It assumes that branching has already occurred at the current depth and reduces 
redundant search by merging semantically similar child nodes via embedding-based clustering. 
In contrast, CiT decides whether branching should occur in the first place by estimating 
the Branching Necessity (BN) at the current node, potentially skipping the entire expansion 
and continuing with a single chained trajectory.}

\red{A second direction focuses on improving \emph{intra-step efficiency} under mandatory branching through neural-network-level optimizations,
such as speculative execution \citep{wang-etal-2025-seed} and prefix-aware tree attention \citep{yao2025deft}.
These methods primarily target reductions in GPU memory footprint and execution overhead at the expansion step, which is not the focus of the first direction (see Appendix~\ref{app:eval_metrics} for details).}

\begin{algorithm*}
\caption{Chain-in-Tree (CiT) with Branching Necessity Evaluation}
\label{alg:continuation}
\small
\textbf{Input}: Depth limit $L$, 
BN budget $k_{\mathrm{bn}}$, thresholds $R_{\mathrm{bn}}, R_{\mathrm{conf}}$, \\ 
LLM roles: Policy $\text{LLM}_{\mathrm{policy}}$,
Transition $\text{LLM}_{\mathrm{trans}}$, Aggregator $\text{LLM}_{\mathrm{agg}}$, Equivalence $\text{LLM}_{\mathrm{eq}}$, BN evaluator $\text{LLM}_{\mathrm{bn}}$ \\
\textbf{Output}: Continued trajectory ending at a node that either triggers branching or is terminal

\vspace{0.2cm}
\noindent
\begin{minipage}[t]{0.48\textwidth}
    \begin{algorithmic}[1]
    \Procedure{BN-DP-Eval}{$s_t, a_t$} 
        \State $r_{\mathrm{bn}} \gets \text{LLM}_{\mathrm{bn}}(s_t, a_t)$
        \State \textbf{return} $r_{\mathrm{bn}}$
    \EndProcedure
    \vspace{0.3cm}
    \Procedure{BN-SC$^1$-Eval}{$s_t, \mathcal{A}_t$} 
        \State $\mathcal{C} \gets \text{LLM}_{\mathrm{agg}}(\mathcal{A}_t)$ \Comment{Cluster candidates}
        \State $r_{\mathrm{bn}} \gets \max_{C \in \mathcal{C}} \frac{|C|}{|\mathcal{A}_t|}$ 
        \State \textbf{return} $r_{\mathrm{bn}}$
    \EndProcedure
    \vspace{0.3cm}
    \Procedure{BN-SC$^2$-Eval}{$s_t, \mathcal{A}_t$} 
        \State $\mathcal{P} \gets \text{pairs from } \mathcal{A}_t$
        \State $\mathcal{C} \gets \{ \text{LLM}_{\mathrm{eq}}(a_i, a_j) : (a_i,a_j) \in \mathcal{P}\}$ \Comment{Equivalence checks}
        \State $r_{\mathrm{bn}} \gets \max_{C \in \mathcal{C}} \frac{|C|}{|\mathcal{A}_t|}$ 
        \State \textbf{return} $r_{\mathrm{bn}}$
    \EndProcedure
    \end{algorithmic}
\end{minipage}%
\hfill
\begin{minipage}[t]{0.48\textwidth}
    \begin{algorithmic}[1]
    \Procedure{CiT-Chain}{$nd, R_{\mathrm{bn}}, R_{\mathrm{conf}}$}
        \State initialize path $\gets [\,]$
        \While{$nd$ is not terminal \red{and $nd.depth \leq L$}}
            \State $\mathcal{A}_t \sim \text{LLM}_{\mathrm{policy}}(nd.state, n=k_{\mathrm{bn}})$
            \State $r_{\mathrm{bn}} \gets \text{BN-Eval}(nd.state, \mathcal{A}_t)$
            \State append $nd$ to path
            \If{$r_{\mathrm{bn}} < R_{\mathrm{bn}}$} 
                \State \textbf{return} path \Comment{branching required}
            \EndIf
            \State $(s', r_{\mathrm{conf}}) \sim \text{LLM}_{\mathrm{trans}}(nd.state, a)$
            \If{$r_{\mathrm{conf}} < R_{\mathrm{conf}}$}
                \State \textbf{return} path \Comment{low transition confidence}
            \EndIf
            \State $nd \gets \langle s', a \rangle$
        \EndWhile
        \State append $nd$ to path
        \State \textbf{return} path
    \EndProcedure
    \end{algorithmic}
\end{minipage}
\end{algorithm*}
\section{Chain-in-Tree}
We introduce \textit{Chain-in-Tree (CiT)}, a plug-and-play chaining phase inserted before tree expansion, i.e., before actually growing the search tree by attaching $k$ new children at the next depth. \footnote{While some search procedures (e.g., MCTS rollouts) may involve branching in the sense of simulating multiple continuations, these branches are not materialized in the tree.}
CiT adaptively decides whether to grow the tree structure versus to continue in-chain, thereby reducing unnecessary expansions.

We finally analyze the efficiency of \textit{Chain-in-Tree} (CiT) under both Tree-of-Thoughts Beam Search (ToT-BS) and MCTS variants.

\subsection{Chaining Phase}
As shown in Algorithm~\ref{alg:continuation}, during the chaining phase, multiple nodes (actions and their resulting states) are generated and concatenated into a linear chain rather than expanded into branches. This occurs when the BN score $r_{\mathrm{bn}}$ exceeds a threshold $R_{\mathrm{bn}}$ and the confidence score $r_{\mathrm{conf}}$ is below a threshold $R_{\mathrm{conf}}$.  

\paragraph{Reusing Children.}
In standard LLM-in-the-loop tree-search (LITS) frameworks, the expansion phase always invokes the policy to generate a full set of $k_{\mathrm{expand}}$ children, regardless of whether a node already has children.  
CiT modifies this behavior: children generated during chaining are reused at expansion time, truncated to at most $k_{\mathrm{expand}}$, and supplemented with new actions from the policy only if fewer than $k_{\mathrm{expand}}$ are available. This design is important to guarantee the efficiency of CiT methods, as specified in Section~\ref{sec:theoretical_bs}and \ref{sec:theoretical_mcts}.
Besides, all children—whether reused or newly generated—are consistently assigned rewards, which are not always required during chaining.  

Concretely:  
1) If $r_{\mathrm{bn}} < R_{\mathrm{bn}}$, the tree is expanded by committing new nodes that contain only the actions produced during chaining; 
2) If $r_{\mathrm{bn}} \geq R_{\mathrm{bn}}$ and $r_{\mathrm{conf}} < R_{\mathrm{conf}}$, both actions and their resulting states are retained for expansion.   

\subsection{Branching Necessity (BN) Evaluation}
\label{sec:bn_eval}
We now describe how BN scores $r_{\mathrm{bn}}$ are computed. We propose two methods: \textbf{Direct Prompting (DP)} and \textbf{Self-Consistency (SC)}.  
In both cases, the evaluation operates over $k_{\mathrm{bn}}$ candidate actions sampled from the policy model $\text{LLM}_{\mathrm{policy}}$. By default, $k_{\mathrm{bn}}=1$ for BN-DP and $k_{\mathrm{expand}}$ for BN-SC.  

\paragraph{Direct Prompting (DP).}
An auxiliary evaluator model $\text{LLM}_{\mathrm{bn}}$ is prompted to directly judge whether an action $a_t$ should trigger branching given the current state $s_t$.  

\paragraph{Self-Consistency (SC).}
A self-consistency strategy \citep{wang2023selfconsistency} is adopted by leveraging the diversity of multiple policy samples. From a state $s_t$, the policy generates a set of $k_{\mathrm{bn}}$ candidate actions $\mathcal{A}_t = \{a_1, \dots, a_{k_{\mathrm{bn}}}\}$.  
Candidates are clustered into equivalence classes.  
The BN score $r_{\mathrm{bn}}$ is then defined as the fraction of actions belonging to the largest cluster.  

\begin{equation}
    r_{\mathrm{bn}} = \frac{\max_{C \in \mathcal{C}} |C|}{k_{\mathrm{bn}}}, 
\end{equation}
where $\mathcal{C} = \text{Equivalence classes of }\mathcal{A}_t$.

Intuitively, if most candidates agree on the next step, the model is confident and chaining is applied. If candidates diverge, branching is triggered. By default, chaining occurs when $r_{\mathrm{bn}} \geq 0.5$ (i.e., more than half of the actions are consistent).  

\paragraph{Two Implementations of BN-SC}  
We explore two clustering implementations:  
1) \textbf{BN-SC$^1$ (Aggregator-based).} An auxiliary model $\text{LLM}_{\mathrm{agg}}$ clusters candidate actions into dictionaries of the form \{canonical action, count\}.  
2) \textbf{BN-SC$^2$ (Pairwise-Equivalence).} A model $\text{LLM}_{\mathrm{eq}}$ is queried as a binary oracle to decide whether two candidate actions are semantically equivalent.  
The union–find–style merging is the applied: each action maintains a representative index, and pairwise equivalences trigger unions. After processing, clusters are defined by representatives and counts are aggregated. In our case, the ``representative index'' is simply the first surviving index arbitrarily assigned during merging.\footnote{In classical union–find, the root is the unique representative in a hierarchy, while no hierarchy is maintained in our case.}

In deterministic domains (e.g., board games with fixed move sets), $\text{LLM}_{\mathrm{eq}}$ can be replaced by simple programmatic rules. Prompts for $\text{LLM}_{\mathrm{bn}}$, $\text{LLM}_{\mathrm{agg}}$, and $\text{LLM}_{\mathrm{eq}}$ are detailed in Appendix~\ref{app:prompts}.

%



\subsection{Efficiency Guarantee of Beam Search}
\label{sec:theoretical_bs}
While LITS frameworks involve multiple LLM roles—policy, reward, and transition models—the cost of policy invocations dominates overall inference and provides the most consistent analytical basis across frameworks. Supporting details are provided in Appendix~\ref{app:theoretical_costs}, with empirical validation in Section~\ref{sec:analysis}.

\begin{assumption*}[Notation and Setup for ToT-BS (also applies to MCTS)]
Let the branching factor be $k_{\mathrm{expand}} \in \mathbb{N}_{>0}$, and let the depth limit be $D \in \mathbb{N}$.  

When a node $v$ is first visited, its \emph{BN score} determines whether $v$ is classified as \emph{easy}.  
If $v$ is easy, the chaining phase is applied: exactly one child is expanded. This consumes $k_{\mathrm{bn}}$ policy calls (with $k_{\mathrm{bn}} \leq k_{\mathrm{expand}}$), but only a single child is preserved. Chaining then continues forward until a node fails the BN threshold.  
At the stopping node (non-easy), a standard full expansion of $k_{\mathrm{expand}}$ children is performed, requiring $k_{\mathrm{expand}}$ policy calls. 
\end{assumption*}

\paragraph{Expansion cost of ToT-BS.}
Figure~\ref{fig:bfs_bn} compares the number of policy invocations of Original ToT-BS vs ToT-BS with chaining.
Let's assume a homogeneous layer, where every expanded node yields exactly$k_{\mathrm{expand}}$ children (or at least $k_{\mathrm{expand}}$ exist) at every depth.
At depth $t$, the frontier size satisfies 
$\min(B,\,k_\mathrm{expand}^t))$ for all $t \ge 0$.
Define the per-depth cost as the number of LLM invocations for action generation, 
the total expansion cost up to depth $D$ is
\begin{equation}
\begin{split}
C_{\mathrm{bs}}(D) \;=\; \sum_{t=0}^{D-1} k_{\mathrm{expand}}\,\min(B,k_{\mathrm{expand}}^t).
\end{split}
\end{equation}

\paragraph{Guarantee.}
Assume that $D_{C1}$ is the first depth that normal beam expansion resumes.
For all $t<D_{C1}$ we have $k_{\mathrm{bn}} \le k_{\mathrm{expand}} \le k_{\mathrm{expand}}\,\min(B,k_{\mathrm{expand}}^t)=C_t$, and for all $t\ge D_{C1}$,
\begin{equation}
\begin{split}
\min(B,\,k_{\mathrm{expand}}^{\,t-D_{C1}}) \;\le\; \min(B,\,k_{\mathrm{expand}}^t),
\end{split}
\end{equation}
hence $C^{\mathrm{cont}}_t \le C_t$. Summing over all depths shows
\begin{equation}
\begin{split}
C_{\mathrm{bs+chain}}(D) \;\le\; C_{\mathrm{bs}}(D).
\end{split}
\end{equation}
Therefore, chaining never increases the number of policy invocations, and yields a strict reduction whenever some $k_{\mathrm{bn}}<k_{\mathrm{expand}}$.
The detailed proof is demonstrated in Appendix~\ref{app:theoretical_costs}.

\subsection{Efficiency Guarantee of MCTS}
\label{sec:theoretical_mcts}
We only consider policy invocation under epxansion. Although there exist policy calls during the simulation phase, chaining shortens the remaining distance to a terminal node before rollout begins, so the simulation phase makes fewer policy calls on average.

\begin{assumption*}[Notation and Setup for MCTS]
In addition to the notation above, let $N \in \mathbb{N}$ denote the number of MCTS iterations (not fixed).  
Following prior work \citep{zhang2024restmcts, hao-etal-2023-reasoning}, we adopt the \emph{full expansion on first visit} rule: an unexpanded node generates all $k_{\mathrm{expand}}$ actions at once when first expanded.  

Define $E(N)$ as the set of distinct nodes first-visited (i.e., expanded) within $N$ MCTS iterations, and let $E^c(N) \subseteq E(N)$ be the subset of those nodes classified as easy under BN evaluation. 
\end{assumption*}

\paragraph{Expansion cost.}
Under the baseline rule, each new node incurs cost $k_{\mathrm{expand}}$, hence
\begin{equation} \begin{aligned}
C_{\text{mcts}}(N) \;=\; k_{\mathrm{expand}} \, |E(N)|.
\end{aligned} 
\label{eq:c_mcts}
\end{equation}
With chaining, easy nodes expand with at most $k_{\mathrm{bn}}$ calls, while hard nodes expand with $k_{\mathrm{expand}}$, so
\begin{equation}
\begin{split}
C_{\text{mcts+chain}}(N) 
&= k_{\mathrm{expand}} \cdot \bigl(|E(N)| - |E^c(N)|\bigr) \\
&\quad+ k_{\mathrm{bn}} \cdot |E^c(N)|.
\end{split}
\label{eq:c_mcts_chain}
\end{equation}

\paragraph{Guarantee.}
Since $k_{\mathrm{bn}}\leq k_{\mathrm{expand}}$, it follows immediately that
\red{\begin{equation}
\begin{split}
C_{\text{mcts+chain}}(N) \;\leq\; C_{\text{mcts}}(N),
\end{split}
\end{equation}}
with strict inequality whenever at least one easy node is encountered.
Thus adding chaining never increases the number of policy invocations and strictly decreases it in the presence of easy nodes.


\section{Experimental Setup}
\label{sec:exp_setup}

\paragraph{Datasets.}
Prior work shows that test-time scaling is most effective when models already possess the required background knowledge \citep{snell2025scaling}.
Accordingly, we evaluate on mathematical reasoning, a domain where Chain-of-Thought (CoT) prompting provides substantial improvements \citep{sprague2025to}. We select two standard benchmarks: \textbf{GSM8K} and \textbf{Math500} \citep{zhang2024accessing,hendrycks2021measuring}. 
For strict evaluation, we retain only problems whose final answers are single numbers, yielding 316 usable examples out of the original 500. From both benchmarks, we use the first 100 test instances for evaluation. To verify that results are not driven by subset variance, we additionally evaluate on all 316 filtered Math500 instances; the efficiency--accuracy trends remain consistent (Appendix~\ref{app:full_math500}).

\paragraph{Baselines.}
We compare CiT against three representative LLM-in-the-loop tree-search (LITS) frameworks. Each serves as the corresponding upper bound of computational cost for its CiT-augmented variant (details in Appendix~\ref{app:alg_details}):  
\textbf{(1) ToT-BS} \citep{yao2023tree}: Beam search where $\text{LLM}_{\mathrm{policy}}$ generates candidate thoughts (actions) and $\text{LLM}_{\mathrm{rm}}$ evaluates candidates. Transitions are realized by concatenation of actions.
\textbf{(2) RAP} \citep{hao-etal-2023-reasoning}: MCTS with dynamic transitions and reward models. Each node is defined by a sub-question (action) and its predicted answer (for formulating the next state). We follow the original settings: policy with 4-shot prompting, temperature $0.8$ (annealed to $0$ at depth limit), generating three candidate branches at each expansion, and 10 MCTS iterations. Rewards are obtained by profiling LLMs for \emph{usefulness}, augmented with an additional correctness check (Appendix~\ref{app:lmpr}), which we also apply to ToT-BS. $\text{LLM}_{\mathrm{trans}}$ is profiled to answer each sub-question.
\textbf{(3) ReST-MCTS* (abbreviated as ReST)}: This method is an MCTS variant where each node corresponds to a free-form reasoning step (``thought'').  
For the \textit{policy}, we use a modified prompt (Appendix~\ref{app:prompts}) with temperature $0.7$. Other MCTS settings are the same as RAP.
For the \textit{transition model}, we follow the same concatenation-based approach as in ToT-BS.  
For the \textit{evaluator}, we require a Process Reward Model (PRM). The original work introduces a PRM but does not release its trained model. Therefore, we adopt the publicly available PRM from \citet{xiong2024rlhflowmath}, which is finetuned on GSM8K and Math500 training sets, making it a suitable substitute for evaluation in our experiments.  
\textbf{(4) CoT} serves as a lower bound for both efficiency and accuracy. 


\paragraph{Base LLMs.}
Our main experiments use \textbf{Qwen3-32B-AWQ}, the strongest open-source model that fits on a single 40GB GPU at the time of writing. We also evaluate with \textbf{LLaMA3-8B-Instruct} as a small language model (SLM), consistent with prior findings that SLMs more clearly reveal the benefits of inference-time search \citep{anonymous2024mutual,zhang2024restmcts}.  
However, RAP requires a completion-style interface, while chat-formatted models (Qwen3-32B-AWQ, LLaMA3-8B-Instruct) enforce a \texttt{(role, content)} structure incompatible with RAP’s design. Therefore, we use the completion model \textbf{LLaMA3-8B} for RAP. See Appendix~\ref{app:no_chat_in_rap} for details.
The parameters for LLM inference, including GPU specifications, decoding temperatures, and maximum output token limits, are detailed in Appendix~\ref{app:llm_params}.  


\paragraph{Evaluation Settings.}
In total, we consider 14 configurations: 
\begin{itemize}
    \item 3 base-LLM settings (Qwen3-32B for all roles; LLaMA3-8B for all roles; LLaMA3-8B for LITS roles with Qwen3-32B for BN roles)  
    $\times$ 2 frameworks (ReST-MCTS, ToT-BS)  
    $\times$ 2 datasets (GSM8K, Math500) 
    = 12 settings.  
    \item Plus RAP with LLaMA3-8B (all roles) on both datasets = 2 additional settings.  
\end{itemize}
\red{CiT does not require a separate BN model.} The mixed setting (Qwen3-32B for BN roles only) is included to evaluate the impact of BN evaluator quality.

\paragraph{Metrics.} 
Efficiency is measured by:  
1) number of output tokens, 
\red{2) number of invocations,}
3) wall-clock runtime of $\text{LLM}_{\mathrm{policy}}$, and  
4) total runtime.    
Accuracy is measured using a lightweight number-only evaluator adapted from RAP \citep{hao-etal-2023-reasoning}, extended with string-to-number normalization and a fallback verification step (via Qwen3-32B) to ensure robustness in parsing numeric answers embedded in text.  

\red{
\paragraph{Complementary Validation on a Planning Task.}
We further evaluate CiT on the BlocksWorld benchmark as an initial
generalization check beyond mathematical reasoning.
Details are provided in Appendix~\ref{app:blocksworld}.
}

\begin{table*}[ht!]
\centering
\footnotesize
\begin{tabular}{
    >{\raggedright\arraybackslash}p{2.2cm}
    >{\raggedleft\arraybackslash}p{0.85cm} 
    >{\raggedleft\arraybackslash}p{0.85cm} 
    >{\raggedleft\arraybackslash}p{0.85cm} 
    >{\raggedleft\arraybackslash}p{1cm} 
    >{\raggedleft\arraybackslash}p{0.85cm} 
    >{\raggedleft\arraybackslash}p{0.85cm} 
    >{\raggedleft\arraybackslash}p{0.85cm} 
    >{\raggedleft\arraybackslash}p{0.9cm} 
    >{\raggedleft\arraybackslash}p{0.9cm} 
    >{\raggedleft\arraybackslash}p{0.85cm} 
}
\toprule
 & \multicolumn{5}{c}{GSM8K} & \multicolumn{5}{c}{Math500} \\
\cmidrule(r){2-6} \cmidrule(l){7-11}
Method & Out & Inv & Time & Total & Acc & Out & Inv & Time & Total & Acc \\
\midrule

\multicolumn{6}{l}{\textbf{Relative to ToT-BS} \; (baseline Acc: \textbf{0.98}; CoT: 0.96)} & \multicolumn{5}{l}{(baseline Acc: \textbf{0.87}; CoT: 0.79)} \\
\quad \textbf{+BN-SC$^1$}
& 13.1\%\rdn & 11.8\%\rdn & 14.7\%\rdn & 16.0\%\rdn & 0.96
& 28.8\%\rdn & 2.8\%\rdn & 36.7\%\rdn & 32.5\%\rdn & 0.89 \\

\quad \textbf{+BN-SC$^2$}
& 35.2\%\rdn & 37.0\%\rdn & 38.0\%\rdn & 44.4\%\rdn & 0.96
& 62.8\%\rdn & 40.3\%\rdn & 72.8\%\rdn & 71.2\%\rdn & 0.84 \\

\quad \textbf{+BN-DP}
& 78.3\%\rdn & 78.3\%\rdn & 78.5\%\rdn & 77.3\%\rdn & 0.97
& 78.9\%\rdn & 80.0\%\rdn & 78.1\%\rdn & 78.2\%\rdn & 0.86 \\
\midrule

\multicolumn{6}{l}{\textbf{Relative to ReST} \; (baseline Acc: \textbf{0.97}; CoT: 0.96)} & \multicolumn{5}{l}{(baseline Acc: \textbf{0.87}; CoT: 0.79)} \\
\quad \textbf{+BN-SC$^1$}
& 26.7\%\rdn & 27.6\%\rdn & 34.3\%\rdn & 12.1\%\gup & 0.97
& 41.4\%\rdn & 43.7\%\rdn & 37.0\%\rdn & 25.2\%\rdn & 0.84 \\

\quad \textbf{+BN-SC$^2$}
& 41.7\%\rdn & 44.0\%\rdn & 49.0\%\rdn & 16.2\%\rdn & 0.97
& 60.7\%\rdn & 58.0\%\rdn & 58.8\%\rdn & 48.7\%\rdn & 0.85 \\

\quad \textbf{+BN-DP}
& 79.6\%\rdn & 80.1\%\rdn & 81.8\%\rdn & 80.3\%\rdn & 0.97
& 82.4\%\rdn & 84.9\%\rdn & 82.8\%\rdn & 82.4\%\rdn & 0.88 \\
\bottomrule
\end{tabular}
\caption{Percentage cost savings relative to ToT-BS and ReST (Qwen3-32B). 
 \textbf{In} = input tokens (policy),  
\textbf{Out} = output tokens (policy),  
\textbf{Inv} = number of model invocations (policy),  
\textbf{Time} = wall-clock running time in hours (policy),  
\textbf{Time (Total)} = overall LLM running time in hours (LLMs as policy, reward models, transition models and BN evaluators),  
\textbf{Acc} = accuracy. \rdn indicates reductions; \gup indicates overhead.}
\label{tab:rel_change_qwen}
\end{table*}

\begin{table*}[ht!]
\centering
\footnotesize
\begin{tabular}{
    >{\raggedright\arraybackslash}p{2.2cm}
    >{\raggedleft\arraybackslash}p{0.85cm} 
    >{\raggedleft\arraybackslash}p{0.85cm} 
    >{\raggedleft\arraybackslash}p{0.85cm} 
    >{\raggedleft\arraybackslash}p{1cm} 
    >{\raggedleft\arraybackslash}p{0.85cm} 
    >{\raggedleft\arraybackslash}p{0.85cm} 
    >{\raggedleft\arraybackslash}p{0.85cm} 
    >{\raggedleft\arraybackslash}p{0.9cm} 
    >{\raggedleft\arraybackslash}p{0.9cm} 
    >{\raggedleft\arraybackslash}p{0.85cm} 
}
\toprule
 & \multicolumn{5}{c}{GSM8K} & \multicolumn{5}{c}{Math500} \\
\cmidrule(r){2-6} \cmidrule(l){7-11}
Method & Out & Inv & Time & Total & Acc & Out & Inv & Time & Total & Acc \\
\midrule
\multicolumn{6}{l}{\textbf{Relative to ToT-BS} \; (baseline Acc: \textbf{0.79}; CoT: 0.68)} & \multicolumn{5}{l}{(baseline Acc: \textbf{0.39}; CoT: 0.34)} \\
\quad \textbf{BN-SC$^1$-}
& 14.9\%\gup & 17.4\%\gup & 13.5\%\gup & 21.6\%\rdn & 0.71
& 34.7\%\gup & 37.7\%\gup & 36.7\%\gup & 25.5\%\gup & 0.35 \\

\quad \textbf{BN-SC$^2$-}
& 22.4\%\rdn & 22.4\%\rdn & 20.2\%\rdn & 80.9\%\rdn & 0.64
& 38.6\%\rdn & 24.8\%\rdn & 38.6\%\rdn & 76.6\%\rdn & 0.30 \\

\quad \textbf{BN-DP-}
& 68.8\%\rdn & 71.0\%\rdn & 68.5\%\rdn & 73.4\%\rdn & 0.73
& 69.9\%\rdn & 63.4\%\rdn & 69.7\%\rdn & 68.8\%\rdn & 0.27 \\

\quad \textbf{BN-SC$^1$+}
& 2.9\%\rdn  & 1.6\%\gup  & 1.1\%\rdn  & 2.3\%\gup  & 0.80
& 36.9\%\gup & 25.5\%\gup & 40.4\%\gup & 62.9\%\gup & 0.38 \\

\quad \textbf{BN-SC$^2$+}
& 29.2\%\rdn & 29.9\%\rdn & 28.1\%\rdn & 72.8\%\rdn & 0.76
& 81.2\%\gup & 25.1\%\gup & 87.5\%\gup & 4.9\%\rdn & 0.39 \\

\quad \textbf{BN-DP+}
& 64.0\%\rdn & 69.5\%\rdn & 62.9\%\rdn & 70.9\%\rdn & 0.77
& 37.8\%\rdn & 37.8\%\rdn & 37.0\%\rdn & 39.9\%\rdn & 0.36 \\
\midrule

\multicolumn{6}{l}{\textbf{Relative to ReST} \; (baseline Acc: \textbf{0.80}; CoT: 0.68)} & \multicolumn{5}{l}{(baseline Acc: \textbf{0.39}; CoT: 0.34)} \\
\quad \textbf{+BN-SC$^1$-}
& 22.8\%\rdn & 26.0\%\rdn & 21.7\%\rdn & 35.4\%\rdn & 0.70
& 23.3\%\rdn & 17.5\%\rdn & 22.4\%\rdn & 19.6\%\rdn & 0.33 \\

\quad \textbf{+BN-SC$^2$-}
& 47.9\%\rdn & 47.5\%\rdn & 47.2\%\rdn & 69.3\%\rdn & 0.73
& 62.8\%\rdn & 51.1\%\rdn & 63.4\%\rdn & 71.0\%\rdn & 0.29 \\

\quad \textbf{+BN-DP-}
& 77.5\%\rdn & 77.8\%\rdn & 76.4\%\rdn & 74.8\%\rdn & 0.68
& 82.4\%\rdn & 81.6\%\rdn & 82.4\%\rdn & 80.6\%\rdn & 0.36 \\

\quad \textbf{+BN-SC$^1$+}
& 41.0\%\rdn & 40.8\%\rdn & 40.6\%\rdn & 36.1\%\rdn & 0.84
& 65.5\%\rdn & 56.7\%\rdn & 66.5\%\rdn & 50.5\%\rdn & 0.43 \\

\quad \textbf{+BN-SC$^2$+}
& 50.2\%\rdn & 52.1\%\rdn & 49.1\%\rdn & 67.6\%\rdn & 0.80
& 39.8\%\rdn & 48.4\%\rdn & 40.9\%\rdn & 53.8\%\rdn & 0.34 \\

\quad \textbf{+BN-DP+}
& 78.8\%\rdn & 80.6\%\rdn & 78.3\%\rdn & 76.3\%\rdn & 0.76
& 66.7\%\rdn & 67.7\%\rdn & 67.1\%\rdn & 65.0\%\rdn & 0.37 \\
\bottomrule
\end{tabular}
\caption{Percentage cost savings relative to ToT-BS and ReST (LLaMA3-8B Instruct), under both \emph{Poor BN} evaluation (–, using LLaMA-3-8B Instruct) and \emph{Accurate BN} evaluation (+, using Qwen-3-32B). }
\label{tab:rel_change_llama}
\end{table*}

\begin{table*}[ht!]
\centering
\footnotesize
\begin{tabular}{
    >{\raggedright\arraybackslash}p{2.2cm}
    >{\raggedleft\arraybackslash}p{0.85cm} 
    >{\raggedleft\arraybackslash}p{0.85cm} 
    >{\raggedleft\arraybackslash}p{0.85cm} 
    >{\raggedleft\arraybackslash}p{1cm} 
    >{\raggedleft\arraybackslash}p{0.85cm} 
    >{\raggedleft\arraybackslash}p{0.85cm} 
    >{\raggedleft\arraybackslash}p{0.85cm} 
    >{\raggedleft\arraybackslash}p{0.9cm} 
    >{\raggedleft\arraybackslash}p{0.9cm} 
    >{\raggedleft\arraybackslash}p{0.85cm} 
}
\toprule
 & \multicolumn{5}{c}{GSM8K} & \multicolumn{5}{c}{Math500} \\
\cmidrule(r){2-6} \cmidrule(l){7-11}
Method & Out & Inv & Time & Total & Acc & Out & Inv & Time & Total & Acc \\
\midrule

\multicolumn{6}{l}{\textbf{Relative to RAP} \; (baseline Acc: \textbf{0.61}; CoT: 0.32)} & \multicolumn{5}{l}{(baseline Acc: \textbf{0.18}; CoT: 0.17)} \\
\quad \textbf{BN-SC$^1$+}
& 65.3\%\rdn & 69.9\%\rdn & 65.2\%\rdn & 47.9\%\rdn & 0.48
& 19.5\%\rdn & 34.5\%\rdn & 9.0\%\rdn & 2.4\%\rdn & 0.27 \\

\quad \textbf{BN-SC$^2$+}
& 78.0\%\rdn & 75.5\%\rdn & 78.3\%\rdn & 80.8\%\rdn & 0.46
& 63.4\%\rdn & 59.6\%\rdn & 61.2\%\rdn & 60.4\%\rdn & 0.21 \\

\quad \textbf{BN-DP+}
& 88.6\%\rdn & 86.4\%\rdn & 88.4\%\rdn & 77.4\%\rdn & 0.57
& 79.1\%\rdn & 80.5\%\rdn & 79.1\%\rdn & 60.8\%\rdn & 0.26 \\
\bottomrule
\end{tabular}
\caption{Percentage cost savings relative to RAP (LLaMA3-8B), where \emph{Accurate BN} evaluation (+) is performed by using Qwen-3-32B for BN roles.}
\label{tab:rel_change_rap}

\end{table*}

\section{Analysis}
\label{sec:analysis}
We report percentage cost savings in Table~\ref{tab:rel_change_qwen}, Table~\ref{tab:rel_change_llama} (for ToT-BS and ReST), and Table~\ref{tab:rel_change_rap} (for RAP).  
The complete set of raw measurements
underlying these relative improvements are provided in Appendix~\ref{app:raw_tables}.

\paragraph{Overall Efficiency and Stability of BN Methods.}
Across all settings (or tables), \textbf{BN-DP} consistently achieves the efficiency guarantee predicted by our theoretical analysis. In particular, BN-DP reduces total runtime by 75--85\% relative to the baselines, while maintaining accuracy comparable to the underlying LITS framework. This makes BN-DP the most reliable and stable choice among our BN evaluation methods.

By contrast, \textbf{BN-SC$^1$} performs well in most configurations but fails in 4 out of 14 settings. All failures occur within the ToT-BS framework \red{(\textbf{Result Rows 1 \& 4} in Table~\ref{tab:rel_change_llama})}: 
GSM8K and Math500 with both policy and BN roles handled by LLaMA (LLaMA+LLaMA), and GSM8K and Math500 with LLaMA as policy and Qwen as the BN evaluator (LLaMA+Qwen).  

\textbf{BN-SC$^2$} is more stable, failing in only 1 out of 14 settings, occurring in the ToT-BS framework on Math500 once with LLaMA+Qwen (\red{\textbf{Result Row 5} } in  Table~\ref{tab:rel_change_llama}).  
\red{Hence, we recommend BN-DP as the default, empirically stable choice for applying CiT.}

\paragraph{Analysis of Failure Settings.}
To better understand the observed failures, we conducted instance-level analyses for all four failure cases of \textbf{BN-SC$^1$} and the failure case of \textbf{BN-SC$^2$}.  
Two consistent patterns emerge: 
1) BN methods remain more efficient than the baseline on most instances---for example, 
on 73\% and 83\% of Math500 cases with LLaMA+QWen under BN-SC$^1$ and BN-SC$^2$, respectively; 
2) aggregate inefficiency is driven by the remaining small subset of instances, where BN methods incur more calls than the baseline.  
This pattern is consistent across all failure settings (Appendix~\ref{app:failure}).
\red{We further analyze failure cases by correlating efficiency regressions with reasoning tree structure (Appendix~\ref{app:cit_regression}). We find that unusually deep or wide trajectories---rather than problem difficulty---are strong predictors of overhead, suggesting that early detection of such patterns could enable dynamic fallback strategies (e.g., switching to BN-DP). We leave the design of such online mitigation mechanisms to future work.}

\begin{figure*}[t!]
    \centering

    \begin{subfigure}{0.45\linewidth}
        
        \includegraphics[width=\linewidth]{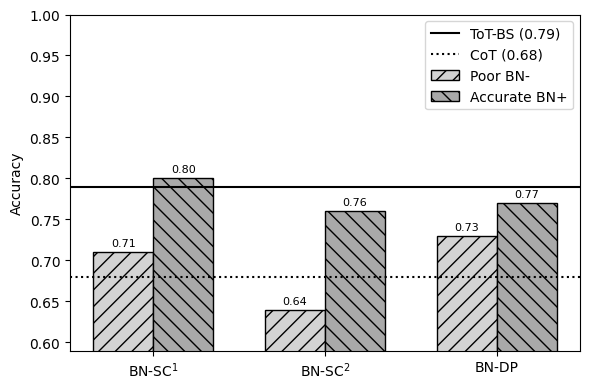}
        \caption{ToT-BS + CiT on GSM8K}
    \end{subfigure}
    \hfill
    \begin{subfigure}{0.45\linewidth}
        
        \includegraphics[width=\linewidth]{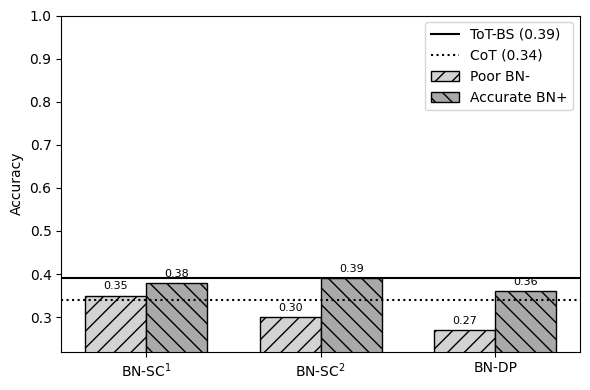}
        \caption{ToT-BS + CiT on Math500}
    \end{subfigure}
    
    \vspace{0.2em}
    
    \begin{subfigure}{0.45\linewidth}
        
        \includegraphics[width=\linewidth]{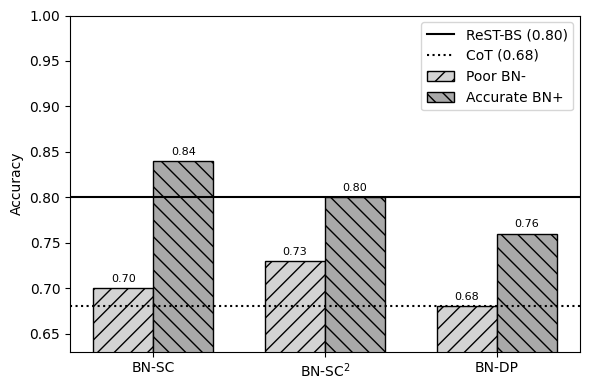}  
        \caption{ReST-MCTS* + CiT on GSM8K}
    \end{subfigure}
    \hfill
    \begin{subfigure}{0.45\linewidth}
        
        \includegraphics[width=\linewidth]{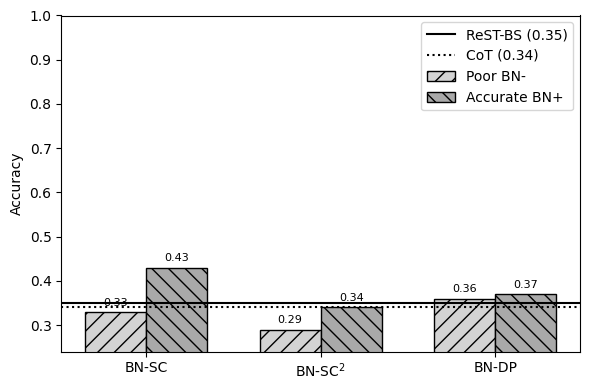}
        \caption{ReST-MCTS* + CiT on Math500}
    \end{subfigure}
        \caption{Accuracy comparison under CiT plug-in across two search frameworks (ReST-MCTS* and ToT-BS) and two datasets (GSM8K, Math500). Bars show BN evaluator quality (\textbf{Poor BN}: LLaMA-3-8B vs \textbf{Accurate BN}: Qwen-3-32B). Horizontal lines denote baselines (CoT, ReST, ToT-BS).}
    \label{fig:poor_vs_accurate_bn}
\end{figure*}

\paragraph{Impact of BN Evaluator Quality on Accuracy.}
Figure~\ref{fig:poor_vs_accurate_bn} illustrates that the quality of BN evaluation plays a critical role in the effectiveness of our CiT plug-in. When smaller LLMs (e.g., LLaMA-3-8B) are used as BN evaluators (\emph{Poor BN}), the accuracy often drops below that of the baseline LITS methods and in some cases approaches the weaker CoT baseline. This suggests that underpowered BN evaluators may systematically overestimate node scores, leading to premature chaining and insufficient exploration, which propagates errors downstream. 

In contrast, when stronger LLMs (e.g., Qwen-3-32B) are used as BN evaluators (\emph{Accurate BN}), the performance recovers and in most cases approaches that of the baseline LITS methods. While in some settings with BN-SC$^1$ performance can even appear higher than baseline LITS, we caution against overinterpreting this result: it may partly reflect the ability of the BN aggregator to reformulate or improve upon actions generated by the policy rather than simply. Thus, we do not claim that BN-SC can outperform baseline LITS. 

The consistent finding across both frameworks (ReST-MCTS* and ToT-BS) is that chaining can be effective when BN evaluation is sufficiently accurate, while poor BN evaluation can negate the benefits of chaining or even harm performance. A reasonable implication is that, for tasks with deterministic action spaces, reliable BN evaluation could in principle be implemented with rule-based or hard-coded checks, thereby eliminating the risks associated with poor BN evaluators.

\paragraph{\red{Additional Planning Evaluation (BlocksWorld).}}
\red{On the BlocksWorld planning task, CiT exhibits efficiency improvements
consistent with those observed on mathematical reasoning benchmarks.
When equipped with BN-DP, CiT preserves a substantial portion of RAP’s planning
performance while reducing inference cost, whereas BN-SC achieves larger
efficiency gains at the expense of solution quality by aggressively suppressing
branching.
Full results are reported in
Appendix~\ref{app:blocksworld}.}

\section{Conclusion}
\label{sec:conclusion}
We proposed \textbf{Chain-in-Tree (CiT)}, a plug-and-play framework that inserts a chaining phase into LLM tree search to avoid unnecessary branching.  
CiT theoretically guarantees non-increasing policy cost and achieves up to 85\% runtime reduction across ToT-BS, ReST-MCTS, and RAP without accuracy loss.  
Overall, CiT emphasizes the importance of accurate BN evaluation for scaling LLM-based search.

\subsection{Reproducibility}
We release an open-source Python package that modularizes LLM-profiled roles across
search frameworks and provides scripts to reproduce all experiments.
The CiT chaining phase is implemented as a single framework-agnostic function.
Implementation details are provided in Appendix~\ref{app:package_design}.

For transparency, the supplementary material includes datasets, execution logs,
search-tree reconstructions, and per-instance inference cost reports.


\section*{Limitations}
\label{limitation}
\paragraph{Coverage of LITS Frameworks.}
Our study \red{focuses on
heuristic-light LITS settings where LLMs drive search decisions.}Other search paradigms such as A* or heuristic-guided search are not evaluated. Nevertheless, our use of unified tasks and LLM-profiled roles \citep{li2025a}, together with modularized implementations, makes it straightforward to extend CiT to additional LITS frameworks in future work.

\paragraph{Scope of Empirical Evaluation.}
Following prior work \citep{snell2025scaling,zhang2024restmcts}, our main experiments focus on mathematical reasoning, where LLMs already possess the necessary knowledge but still benefit from test-time scaling.  
We include BlocksWorld (Appendix~\ref{app:blocksworld}) as an initial evaluation on a planning domain with deterministic action spaces, but broader coverage of such domains (e.g., navigation, other board games) where BN-SC's use of LLMs for clustering semantically equivalent actions may be unnecessary remains future work.

 \paragraph{Accuracy Gains from CiT.}
Although CiT is designed for efficiency, we observe that it can also improve accuracy in several settings. Understanding why chaining sometimes enhances reasoning quality requires deeper analysis, which we leave for future investigation.

\bibliography{custom}

@inproceedings{wang-etal-2025-seed,
    title = "{SEED}: Accelerating Reasoning Tree Construction via Scheduled Speculative Decoding",
    author = "Wang, Zhenglin  and
      Wu, Jialong  and
      Lai, Yilong  and
      Zhang, Congzhi  and
      Zhou, Deyu",
    editor = "Rambow, Owen  and
      Wanner, Leo  and
      Apidianaki, Marianna  and
      Al-Khalifa, Hend  and
      Eugenio, Barbara Di  and
      Schockaert, Steven",
    booktitle = "Proceedings of the 31st International Conference on Computational Linguistics",
    month = jan,
    year = "2025",
    address = "Abu Dhabi, UAE",
    publisher = "Association for Computational Linguistics",
    url = "https://aclanthology.org/2025.coling-main.328/",
    pages = "4920--4937",
}

@inproceedings{wang-etal-2025-dont,
    title = "Don{'}t Get Lost in the Trees: Streamlining {LLM} Reasoning by Overcoming Tree Search Exploration Pitfalls",
    author = "Wang, Ante  and
      Song, Linfeng  and
      Tian, Ye  and
      Yu, Dian  and
      Mi, Haitao  and
      Duan, Xiangyu  and
      Tu, Zhaopeng  and
      Su, Jinsong  and
      Yu, Dong",
    editor = "Che, Wanxiang  and
      Nabende, Joyce  and
      Shutova, Ekaterina  and
      Pilehvar, Mohammad Taher",
    booktitle = "Proceedings of the 63rd Annual Meeting of the Association for Computational Linguistics (Volume 1: Long Papers)",
    month = jul,
    year = "2025",
    address = "Vienna, Austria",
    publisher = "Association for Computational Linguistics",
    url = "https://aclanthology.org/2025.acl-long.1167/",
    doi = "10.18653/v1/2025.acl-long.1167",
    pages = "23946--23959",
    ISBN = "979-8-89176-251-0",
}

@article{sutskever2014sequence,
  title={Sequence to sequence learning with neural networks},
  author={Sutskever, Ilya and Vinyals, Oriol and Le, Quoc V},
  journal={Advances in neural information processing systems},
  volume={27},
  year={2014}
}

@misc{xiong2024rlhflowmath,
      author={Wei Xiong and Hanning Zhang and Nan Jiang and Tong Zhang},
  title = {An Implementation of Generative PRM},
  year = {2024},
  publisher = {GitHub},
  journal = {GitHub repository},
  howpublished = {\url{https://github.com/RLHFlow/RLHF-Reward-Modeling}}
}

@article{li2025entropy,
  title={Entropy-Aware Branching for Improved Mathematical Reasoning},
  author={Li, Xianzhi and Callanan, Ethan and Zhu, Xiaodan and Sibue, Mathieu and Papadimitriou, Antony and Mahfouz, Mahmoud and Ma, Zhiqiang and Liu, Xiaomo},
  journal={arXiv preprint arXiv:2503.21961},
  year={2025}
}

@inproceedings{guez2012lazysampling,
author = {Guez, Arthur and Silver, David and Dayan, Peter},
title = {Efficient Bayes-adaptive reinforcement learning using sample-based search},
year = {2012},
publisher = {Curran Associates Inc.},
address = {Red Hook, NY, USA},
booktitle = {Proceedings of the 26th International Conference on Neural Information Processing Systems - Volume 1},
pages = {1025–1033},
numpages = {9},
location = {Lake Tahoe, Nevada},
series = {NIPS'12}
}

@article{
li2025a,
title={A Survey on {LLM} Test-Time Compute via Search: Tasks, {LLM} Profiling, Search Algorithms, and Relevant Frameworks},
author={Xinzhe Li},
journal={Transactions on Machine Learning Research},
issn={2835-8856},
year={2025},
url={https://openreview.net/forum?id=x9VQFjtOPS},
note={}
}

@misc{
dai2025process,
title={Process Supervision-Guided Policy Optimization for Code Generation},
author={Ning Dai and Zheng Wu and Renjie Zheng and Ziyun Wei and Wenlei Shi and Xing Jin and Guanlin Liu and Chen Dun and Liang Huang and Lin Yan},
year={2025},
url={https://openreview.net/forum?id=Cn5Z0MUPZT}
}

@misc{
luo2025improve,
title={Improve Mathematical Reasoning in Language Models with Automated Process Supervision},
author={Liangchen Luo and Yinxiao Liu and Rosanne Liu and Samrat Phatale and Meiqi Guo and Harsh Lara and Yunxuan Li and Lei Shu and Lei Meng and Jiao Sun and Abhinav Rastogi},
year={2025},
url={https://openreview.net/forum?id=KwPUQOQIKt}
}

@article{zhang2024accessing,
  title={Accessing gpt-4 level mathematical olympiad solutions via monte carlo tree self-refine with llama-3 8b},
  author={Zhang, Di and Huang, Xiaoshui and Zhou, Dongzhan and Li, Yuqiang and Ouyang, Wanli},
  journal={arXiv preprint arXiv:2406.07394},
  year={2024}
}

@inproceedings{
sprague2025to,
title={To CoT or not to CoT? Chain-of-thought helps mainly on math and symbolic reasoning},
author={Zayne Rea Sprague and Fangcong Yin and Juan Diego Rodriguez and Dongwei Jiang and Manya Wadhwa and Prasann Singhal and Xinyu Zhao and Xi Ye and Kyle Mahowald and Greg Durrett},
booktitle={The Thirteenth International Conference on Learning Representations},
year={2025},
url={https://openreview.net/forum?id=w6nlcS8Kkn}
}

@inproceedings{
anonymous2024mutual,
title={Mutual Reasoning Makes Smaller {LLM}s Stronger Problem-Solver},
author={Qi, Zhenting and Ma, Mingyuan and Xu, Jiahang and Zhang, Li Lyna and Yang, Fan and Yang, Mao},
booktitle={The Thirteenth International Conference on Learning Representations},
year={2025},
url={https://openreview.net/forum?id=6aHUmotXaw}
}

@inproceedings{
snell2025scaling,
title={Scaling Test-Time Compute Optimally Can be More Effective than Scaling {LLM} Parameters},
author={Charlie Victor Snell and Jaehoon Lee and Kelvin Xu and Aviral Kumar},
booktitle={The Thirteenth International Conference on Learning Representations},
year={2025},
url={https://openreview.net/forum?id=4FWAwZtd2n}
}

@inproceedings{
yao2025deft,
title={De{FT}: Decoding with Flash Tree-attention for Efficient Tree-structured {LLM} Inference},
author={Jinwei Yao and Kaiqi Chen and Kexun Zhang and Jiaxuan You and Binhang Yuan and Zeke Wang and Tao Lin},
booktitle={The Thirteenth International Conference on Learning Representations},
year={2025},
url={https://openreview.net/forum?id=2c7pfOqu9k}
}

@inproceedings{
zhang2024restmcts,
title={Re{ST}-{MCTS}*: {LLM} Self-Training via Process Reward Guided Tree Search},
author={Dan Zhang and Sining Zhoubian and Ziniu Hu and Yisong Yue and Yuxiao Dong and Jie Tang},
booktitle={The Thirty-eighth Annual Conference on Neural Information Processing Systems},
year={2024},
url={https://openreview.net/forum?id=8rcFOqEud5}
}

@inproceedings{chen-etal-2024-tree,
    title = "When is Tree Search Useful for {LLM} Planning? It Depends on the Discriminator",
    author = "Chen, Ziru  and
      White, Michael  and
      Mooney, Ray  and
      Payani, Ali  and
      Su, Yu  and
      Sun, Huan",
    editor = "Ku, Lun-Wei  and
      Martins, Andre  and
      Srikumar, Vivek",
    booktitle = "Proceedings of the 62nd Annual Meeting of the Association for Computational Linguistics (Volume 1: Long Papers)",
    month = aug,
    year = "2024",
    address = "Bangkok, Thailand",
    publisher = "Association for Computational Linguistics",
    url = "https://aclanthology.org/2024.acl-long.738",
    doi = "10.18653/v1/2024.acl-long.738",
    pages = "13659--13678",
}

@inproceedings{
hendrycks2021measuring,
title={Measuring Mathematical Problem Solving With the {MATH} Dataset},
author={Dan Hendrycks and Collin Burns and Saurav Kadavath and Akul Arora and Steven Basart and Eric Tang and Dawn Song and Jacob Steinhardt},
booktitle={Thirty-fifth Conference on Neural Information Processing Systems Datasets and Benchmarks Track (Round 2)},
year={2021},
url={https://openreview.net/forum?id=7Bywt2mQsCe}
}

@misc{
zhou2024language,
title={Language Agent Tree Search Unifies Reasoning Acting and Planning in Language Models},
author={Andy Zhou and Kai Yan and Michal Shlapentokh-Rothman and Haohan Wang and Yu-Xiong Wang},
year={2024},
url={https://openreview.net/forum?id=6LNTSrJjBe}
}

@inproceedings{
valmeekam2023planbench,
title={PlanBench: An Extensible Benchmark for Evaluating Large Language Models on Planning and Reasoning about Change},
author={Karthik Valmeekam and Matthew Marquez and Alberto Olmo and Sarath Sreedharan and Subbarao Kambhampati},
booktitle={Thirty-seventh Conference on Neural Information Processing Systems Datasets and Benchmarks Track},
year={2023},
url={https://openreview.net/forum?id=YXogl4uQUO}
}

@inproceedings{hao-etal-2023-reasoning,
    title = "Reasoning with Language Model is Planning with World Model",
    author = "Hao, Shibo  and
      Gu, Yi  and
      Ma, Haodi  and
      Hong, Joshua  and
      Wang, Zhen  and
      Wang, Daisy  and
      Hu, Zhiting",
    editor = "Bouamor, Houda  and
      Pino, Juan  and
      Bali, Kalika",
    booktitle = "Proceedings of the 2023 Conference on Empirical Methods in Natural Language Processing",
    month = dec,
    year = "2023",
    address = "Singapore",
    publisher = "Association for Computational Linguistics",
    url = "https://aclanthology.org/2023.emnlp-main.507",
    pages = "8154--8173",
}

@inproceedings{
wang2023selfconsistency,
title={Self-Consistency Improves Chain of Thought Reasoning in Language Models},
author={Xuezhi Wang and Jason Wei and Dale Schuurmans and Quoc V Le and Ed H. Chi and Sharan Narang and Aakanksha Chowdhery and Denny Zhou},
booktitle={The Eleventh International Conference on Learning Representations },
year={2023},
url={https://openreview.net/forum?id=1PL1NIMMrw}
}

@misc{yao2023tree,
      title={Tree of Thoughts: Deliberate Problem Solving with Large Language Models}, 
      author={Shunyu Yao and Dian Yu and Jeffrey Zhao and Izhak Shafran and Thomas L. Griffiths and Yuan Cao and Karthik Narasimhan},
      year={2023},
      eprint={2305.10601},
      archivePrefix={arXiv},
      primaryClass={cs.CL}
}

@inproceedings{yao2023react,
    title={ReAct: Synergizing Reasoning and Acting in Language Models},
    author={Shunyu Yao and Jeffrey Zhao and Dian Yu and Nan Du and Izhak Shafran and Karthik R Narasimhan and Yuan Cao},
    booktitle={The Eleventh International Conference on Learning Representations },
    year={2023},
    url={https://openreview.net/forum?id=WE_vluYUL-X}
}

\appendix

\section{Algorithmic Design of Search Control}
\label{app:novelty_clarification}


\paragraph{Search Control with CiT.}
\red{CiT introduces a new chaining phase that intervenes before node expansion, altering the control flow of tree search by deciding whether structural branching should occur at all. This requires nontrivial algorithmic changes, including: 
(i) reusing partially generated children across chaining and expansion phases; 
(ii) decoupling action generation from reward assignment and state expansion so to save unnecessary inference cost for chaining nodes; 
(iii) hence, during the expansion phase, additional checking is required to ensure rewards for nodes generated during chaining phase but not identified for chaining 
iiii) identifying chaining nodes and treating them correctly within iterative processes and across other search phases, including an additional check before UCT selection. 
These design choices are essential for making a seemingly simple principle work uniformly across ToT-BS, ReST-MCTS, and RAP, and are fundamentally different from prior approaches that operate after expansion (e.g., node merging) or at the token level (e.g., speculative decoding).}

\paragraph{Task-Agnostic Abstraction for Tree Search.}
\red{We generalize the task abstraction to support BN evaluation and chaining decisions that are independent of task formulation (e.g., although action is more direct abstraction for policy. abstracting it as a part of Step can streamline the information flow from Policy to Transition and Reward Models. The task-specific attributes beyond actions, such as observation, state snapshots in BlocksWorld, but are necessary for transition and reward models can be instantiated) and framework-specific prompt structures. This design is essential for making CiT truly plug-and-play across different frameworks, and goes beyond prior work that tightly couples search logic to task-specific representations. Concretely, we unify the original query in language reasoning tasks and the evolving goal into a single state in environment-grounded tasks (e.g., BlocksWorld). Such abstraction can be uniformly consumed by the search procedure with different policy, reward models, and transition model.}

\section{Cost and Resource Analysis}
\label{app:eval_metrics}
\red{
We evaluate efficiency primarily using token counts (input and output), model invocations, and wall-clock runtime. Below we clarify how these metrics relate to other commonly discussed cost measures.
}

\red{
\paragraph{FLOPs.}
For decoder-only Transformer architectures with fixed model size and implementation, the dominant computational cost at inference arises from forward passes over input prefixes and generated tokens. As a result, the total number of input and output tokens provides a close proxy for total FLOPs, up to constant factors determined by the architecture and hardware. This practice is standard in LLM-in-the-loop tree search (LITS) and test-time compute (TTC) literature (e.g., ReST-MCTS*), where token usage and runtime are used to characterize computational efficiency. Consequently, we do not separately report FLOPs.
}

\red{
\paragraph{Memory Footprint.}
All methods compared in this work use the same base model and run on identical hardware configurations. CiT does not modify the underlying model architecture, context length, or execution graph. Therefore, peak GPU memory usage remains effectively constant across baselines and CiT variants in typical (non-low-resource) settings. Accurately measuring fine-grained GPU memory dynamics would require system- and implementation-specific profiling, which is orthogonal to our algorithmic focus and omitted here.
}

\red{
\paragraph{API Cost.}
Our experiments are conducted using local open-weight models. For commercial LLM APIs, pricing is typically defined as a linear function of the number of input and output tokens processed. Under such pricing models, the token reductions achieved by CiT translate directly into proportional reductions in API cost. As API pricing is provider-specific, we do not report absolute monetary costs.
}

\section{Prompts}
\label{app:prompts}
The prompt use as policy, reward model and world model is summarized in Table~\ref{tab:prompt_smmary}.
\begin{table*}[htbp!] 
\centering 

\footnotesize
\begin{tabular}{p{1cm}p{3cm}p{3cm}p{3cm}}
    \toprule
     Methods  & Policy & Reward Model & Dynamic Model 
    \\ \midrule
    RAP
    & QA-based policy (Table~\ref{tab:rap_policy_transition})
    & Usefulness logit-based evaluator (Table~\ref{tab:rap_eval})
    & QA-based transition model (Table~\ref{tab:rap_policy_transition})
    \\ \addlinespace
    
    ReST-MCTS*
    & Thought generator (Table~\ref{tab:policy_rest})
    & Fine-tuned reward models
    & NA
    \\ \addlinespace

    ToT-BF
    & Thought generator (Table~\ref{tab:policy_rest})
    & Usefulness evaluator(Table~\ref{tab:usefulness_eval}) + Correctness evaluator (Table~\ref{tab:correctness_eval})
    & NA
    \\ \bottomrule
\end{tabular}
\caption{Prompt Summary for Different Frameworks.} 
\label{tab:prompt_smmary}
\end{table*}
\begin{table*}[!h]
    \centering
       
    \begin{tabular}{p{12cm}}
    \toprule
        Given a question, please decompose it into sub-questions. For each sub-question, please answer it in a complete sentence, ending with "The answer is". When the original question is answerable, please start the subquestion with "Now we can answer the question:" \\
       \textit{(few-shot demonstrations are ignored)}\\
    \textbf{Question 1:} James writes a 3-page letter to 2 different friends twice a week. How many pages does he write a year? \\
    \textbf{Question 1.1:} How many pages does he write every week? \\
    \textit{Answer 1.1:} James writes a 3-page letter to 2 different friends twice a week, so he writes 3 * 2 * 2 = 12 pages every week. The answer is 12. \\
    \textbf{Question 1.2:} How many weeks are there in a year? \\
    \textit{Answer 1.2:} There are 52 weeks in a year. The answer is 52. \\
    \textbf{Question 1.3:} Now we can answer the question: How many pages does he write a year? \\
    \textit{Answer 1.3:} James writes 12 pages every week, so he writes 12 * 52 = 624 pages a year. The answer is 624. \\
    ... \\ \\
        
         \textbf{Question 5:} Janet’s ducks lay 16 eggs per day. She eats three for breakfast every morning and bakes muffins for her friends every day with four. She sells the remainder at the farmers' market daily for \$2 per fresh duck egg. How much in dollars does she make every day at the farmers' market? \\

        \textbf{Question 5.1:} \begin{myshadowbox} 
        \textit{(Generation as a policy)} 
        \\ How many eggs does Janet have left after eating three for breakfast and baking muffins with four?  \end{myshadowbox} \\

        \textbf{Answer 5.1:} 
        \begin{myshadowbox} 
        \textit{(Generation as a transition model)} \\
        Janet starts with 16 eggs per day, consumes 3 for breakfast and 4 for baking, so she has 16 - 3 - 4 = 9 eggs left for the farmers' market. The answer is 9.
        \end{myshadowbox} \\
      \bottomrule
    \end{tabular}
 \caption{A policy/transition model for RAP. Sourced from \citet{hao-etal-2023-reasoning}.}
    \label{tab:rap_policy_transition}
\end{table*}

\begin{table*}[!h]
    \centering
    
    \begin{tabular}{p{12cm}}
    \toprule
    \textbf{System message:} \\\addlinespace
    Your task is to give the correct next step, given a science problem and an existing partial solution (not a complete answer).  
    \\
    Assuming the input is n-steps, then the format of the input is:\\ "Problem: ...\\ Existing Steps:\\ Step 1: ...\\ Step 2: ...\\ ...\\ Step n: ..."\\ where ... denotes omitted input information. \\ \\ Please follow the restricted output format:\\ * If no existing steps are given, generate Step 1.\\ * Otherwise, following the given step(s), output ONLY ONE step within 1000 tokens. SHOULD NOT output multiple steps.\\ * DO NOT repeat Problem or any Existing Steps.\\ * Your output should be a complete reasoning step that includes calculations, reasoning, choosing answers, etc. \\ * When the final answer is/has been reached, begin the step with EXACTLY the phrase:"Now we can answer the question: The answer is ", followed by EXACTLY one number. Do not include any other words, punctuation, or explanation after the number. \\
    \midrule
    \textbf{User message:} \\\addlinespace
    Problem: How many positive whole-number divisors does 196 have?\\ Existing Steps: None\\ Step 1: \\
    \bottomrule
    \end{tabular}
    \caption{A policy used for ReST and BFS. Sourced from \citet{zhang2024restmcts}.}
    \label{tab:policy_rest}
\end{table*}

\begin{table*}[!h]
    \centering
        
    \begin{tabular}{p{12cm}}
    \toprule
        Given a question and some sub-questions, determine whether the last sub-question is useful to answer the question. Output 'Yes' or 'No', and a reason. \\
        \textbf{Question 1:} Four years ago, Kody was only half as old as Mohamed. If Mohamed is currently twice as 30 years old, how old is Kody? \\
        \textbf{Question 1.1:} How old is Mohamed? \\
        \textbf{Question 1.2:} How old was Mohamed four years ago? \\
        \textbf{New question 1.3:} How old was Kody four years ago? \\
        \textit{Is the new question useful? Yes. We need the answer to calculate how old is Kody now.} \\ 
        \textit{(Few-shot demonstrations are ignored.)} \\\\
        \textbf{Question 5:} Janet’s ducks lay 16 eggs per day. She eats three for breakfast every morning and bakes muffins for her friends every day with four. She sells the remainder at the farmers' market daily for \$2 per fresh duck egg. How much in dollars does she make every day at the farmers' market? \\
        \textbf{New question 5.1:} Now we can answer the question: How much in dollars does she make every day at the farmers' market? \\
        \textit{Is the new question useful? } \\
    \bottomrule
    \end{tabular}
    \caption{An LLM evaluator. Source from \citet{hao-etal-2023-reasoning}}
    \label{tab:rap_eval}
\end{table*}

\begin{table*}[!ht]
    \centering
       
    \begin{tabular}{p{12cm}}
    \toprule
\textbf{System message:} \\\addlinespace
Given a question and a chain of thoughts, determine whether the last thought is **correct**, where **correct** means factually accurate and mathematically accurate (all calculations and formulas are correct), and logically consistent with the question.\\
Instructions:\\
Output only a score:\\
- 0 if any correctness criterion is unmet.\\
- 1 if all correctness criteria are fully met.\\
The score must be parsable by Python's \texttt{float()} function, with no punctuation or additional text. \\
\midrule
\textbf{User message:} \\\addlinespace
Problem: How many positive whole-number divisors does 196 have?\\
Existing Steps:\\
Step 1: Find the prime factorization of 196 as $2^2 \times 7^2$\\
Step 2: Apply the divisor formula $(a+1)(b+1)$ to the prime factorization $2^2 \times 7^2$ and calculate $3 \times 3 = 9$\\
New Step to be evaluated: Now we can answer the question: The answer is 9. 
\\ \bottomrule
    \end{tabular}
     \caption{A correctness evaluator used for BFS.}
    \label{tab:correctness_eval}
\end{table*}

\begin{table*}[!h]
    \centering
       
    \begin{tabular}{p{12cm}}
    \toprule
\textbf{System message:} \\\addlinespace
Given a question and a chain of thoughts, determine how **useful** the last thought is for answering the question, regardless of correctness. \\
Instructions: \\
Output only a score between 0 and 1: \\
- 0 if the step is entirely irrelevant or unhelpful. \\
- 1 if the step is essential and maximally useful. \\
- A value strictly between 0 and 1 if the step is partially useful. Larger values indicate more usefulness. \\
The score must be parsable by Python's \texttt{float()} function, with no punctuation or additional text. \\

\\ \midrule

\textbf{User message:} \\\addlinespace
Problem: How many positive whole-number divisors does 196 have? \\
Existing Steps: \\
Step 1: Find the prime factorization of 196 as $2^2 \times 7^2$ \\
Step 2: Apply the divisor formula $(a+1)(b+1)$ to the prime factorization $2^2 \times 7^2$ and calculate $3 \times 3 = 9$ \\
New Step to be evaluated: Now we can answer the question: The answer is 9. \\
\\ \bottomrule
    \end{tabular}
     \caption{A usefulness evaluator used for BFS.}
    \label{tab:usefulness_eval}
\end{table*}

\begin{table*}[!h]
    \centering
       
    \begin{tabular}{p{12cm}}
    \toprule
\textbf{System message:} \\\addlinespace
You are an expert at deciding whether a single reasoning
step is *logically compulsory* given the task and the partial solution path.\\

Input fields
(A) Task description - one paragraph.\\
(B) Partial reasoning path so far.\\
(C) Candidate next step.\\

ONLY output a single number from 1 to 4.\\

Scale\\
4 - **Unavoidable next step**: given the current path, this step must come next to proceed logically.  \\
3 - Strongly expected: skipping it now would be very unusual, though not impossible.  \\
2 - Potentially useful but avoidable: alternative coherent next steps exist.  \\
1 - **Optional**: the step is not logically required at this point.\\

Think silently, then output the single line - nothing else.

\\ \midrule

\textbf{User message:} \\\addlinespace
(A) \textit{(task)} \\
(B) \textit{(partial path)}\\
(C) \textit{(candidate step)}
\\ \bottomrule
    \end{tabular}
     \caption{BN Evaluator.}
    \label{tab:prompt_bnd}
\end{table*}

\begin{table*}[!h]
    \centering
     
    \begin{tabular}{p{12cm}}
    \toprule
    \textbf{System message for RAP:} \\\addlinespace
You are given a QUESTION and its partial solution (Subquestions which have been answered).\\
Your task is to group the provided list of candidate next subquestions (After "List of Candidates for the following step") into clusters.\\
\\
- Steps that are semantically equivalent must be grouped together.\\
- Paraphrase or stylistic differences are irrelevant\\
- Existing Steps are given only as context and MUST NOT appear in the clusters.\\
\\
OUTPUT FORMAT (Python literal list only; must be parsable by ast.literal\_eval):\\
OUTPUT FORMAT: \\
\texttt{[}\\
\texttt{ \{ "canonical\_action": "<a CONCRETE subquestion>", "count": <the number of the candidates grouped in that cluster> \},}\\
\texttt{ ...}\\
\texttt{]}\\
Rules:\\
- Each array element represents one cluster.\\
- No text outside the list.\\
- The total number of generated words should be NO more than 450 words.\\
\\
    \\ \midrule

    \textbf{System message for ReST-MCTS* and ToT-BS:} \\ \addlinespace
    You are given a QUESTION and its partial solution (Existing Steps).\\
Your task is to group the provided list of candidate next steps (After "List of Candidates for the following step") into clusters.\\
\\
- Steps that are semantically equivalent must be grouped together.\\
- Paraphrase or stylistic differences are irrelevant.\\
- Existing Steps are given only as context and MUST NOT appear in the clusters.\\
\\
\textbf{OUTPUT FORMAT:} \\
\texttt{[}\\
\texttt{ \{ "canonical\_action": "<CONCRETE calculation(s) and outcome(s) after the Existing Steps>", "count": <the number of the candidates grouped in that cluster> \},}\\
\texttt{ ...}\\
\texttt{]}\\
Rules: Each array element represents one cluster. No text outside the list. The total number of generated words should be no more than 450 words.\\
    \\ \bottomrule
    \end{tabular}
       \caption{BN Aggregator .}
    \label{tab:prompt_bn_agg}
\end{table*}

\begin{table*}[!h]
    \centering
   
    \begin{tabular}{p{12cm}}
    \toprule
\textbf{System message for RAP:} \\
You are a strict semantic comparator.\\
Given two sub-questions, decide if they are semantically overlapping given the context.
\\ \addlinespace\midrule \addlinespace
\textbf{System message for ReST-MCTS* and ToT-BS:} \\
You are a strict semantic comparator.\\
Given two action descriptions, decide if they are semantically overlapping given the context.
\\
Definition: \\
- "Overlapping" means the two descriptions express the same underlying operation or one is a specific case/subsumption of the other or have the same effect on the context.\\
- "Not overlapping" means the operations are mutually exclusive in meaning.\\
Answer format: return only 'YES' or 'NO' with no punctuation, no explanation.
\\ \addlinespace\midrule \addlinespace
\textbf{User message:} \\
Context: \\
======== \\
\{\textit{context}\} \\
======== \\
New Step A: \{\textit{canonical}\_\textit{action}\} \\
New Step B: \{\textit{canonical}\_\textit{action}\} \\
Do these steps express the same underlying operation given the context?
\\ \addlinespace\bottomrule 
\end{tabular}
 \caption{BN equivalence checker}
    \label{tab:prompt_bn_eq}
\end{table*}



\clearpage   
\section{Theoretical Costs}
\label{app:theoretical_costs}

\subsection{Scope of Theoretical Analysis}
\label{app:why_not}
This section clarifies why our theoretical efficiency analysis of \textit{Chain-in-Tree} (CiT) focuses exclusively on policy invocations.  


\paragraph{Justification for Focusing on Policy Invocations.}
\textbf{1) \red{Cost of reward and transition models never increase}}:
While LLMs may serve as \emph{policy}, \emph{reward}, and \emph{transition} models, we only count invocations of $\text{LLM}_{\mathrm{policy}}$. The reason is structural: every node generated by $\text{LLM}_{\mathrm{policy}}$ necessarily triggers the use of $\text{LLM}_{\mathrm{rm}}$ and/or $\text{LLM}_{\mathrm{trans}}$ if it exists. For example, in ToT-BS, every child node requires reward evaluation; in RAP, at least one transition evaluation is needed per depth to maintain expansion. 
By contrast, during the chaining phase, CiT invokes $\text{LLM}_{\mathrm{trans}}$ exactly once per chaining node (i.e., per depth), which is no worse than the per-depth transition usage in the original frameworks; moreover, reward evaluation via $\text{LLM}_{\mathrm{rm}}$ can be deferred until branching. Thus, CiT consistently reduces reward/transition overhead under the same policy budget.

\textbf{(2) Additional BN evaluation roles are lightweight.} 
Although CiT introduces additional roles---$\text{LLM}_{\mathrm{bn}}$, $\text{LLM}_{\mathrm{agg}}$, and $\text{LLM}_{\mathrm{eq}}$---for Branching Necessity (BN) evaluation, their overhead is significantly smaller than that of $\text{LLM}_{\mathrm{policy}}$ or $\text{LLM}_{\mathrm{trans}}$ (when present). We confirm this empirically in Section~\ref{sec:analysis}.

\paragraph{Uniform Token Length Assumption.}
We assume the number of input and output tokens per $\text{LLM}_{\mathrm{policy}}$ invocation is approximately uniform across steps and examples. 

\begin{figure*}[!ht]
    \centering
  
    \begin{subfigure}{0.48\textwidth}
        \centering
       \includegraphics[width=\textwidth]{pics/s3/bfs.jpg}
        \caption{ToT-Beam Search.}
        
        \label{fig:bfs_in_app}
    \end{subfigure}
    \hfill
    \begin{subfigure}{0.48\textwidth}
        \centering
       
        \includegraphics[width=\textwidth]{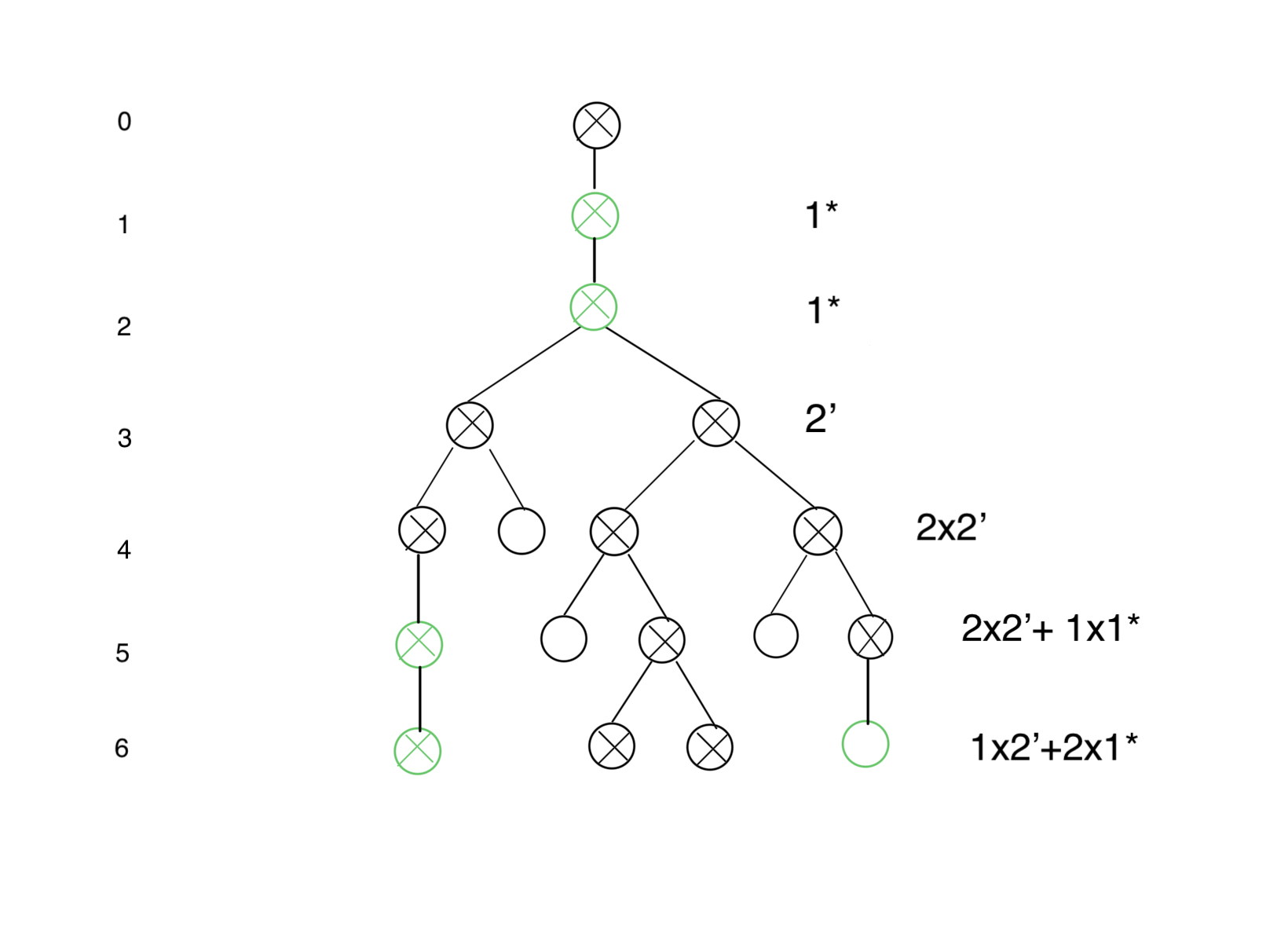}
         \caption{ToT-Beam Search + Chaining (BN-DP).}\label{fig:bfs_bnsc_in_app}
    \end{subfigure}
      \caption{The number of policy invocations of Original ToT-BS vs ToT-BS + Chaining. The numbers suffixed by ``$\prime$'' indicate sampling size for beam search expansion, while The numbers suffixed by ``*'' indicate sampling size for BN judge.}
    \label{fig:bfs_bn_in_app}
\end{figure*}

\subsection{ToT-BS}
\label{app:theoretical_costs_bs}
\proposition{prop:frontier-size}
(Beam-search frontier size.)
At depth $t$, the frontier size under beam search satisfies 
\begin{equation}
\begin{aligned}
|N_t| = \min(B, k_{\mathrm{expand}}^t),
\end{aligned}
\end{equation}
assuming each node generates at most $k_{\mathrm{expand}}$ children, pruning retains the top $B$ nodes at each layer, and no terminals are reached.

\medskip
\noindent \textit{Proof.} 
The recurrence relation for the frontier is
\begin{equation}
\begin{aligned}
|N_{t+1}| = \min\!\bigl(B,\, k_{\mathrm{expand}}\,|N_t|\bigr).
\end{aligned}
\end{equation}
For the base case, $|N_0| = 1 = \min(B,k_{\mathrm{expand}}^0)$ (assuming $B \ge 1$). 
For the inductive step, assume $|N_t| = \min(B, k_{\mathrm{expand}}^t)$. Then

\begin{equation}
\begin{split}
|N_{t+1}| 
&= \min\!\bigl(B,\, k_{\mathrm{expand}} \min(B, k_{\mathrm{expand}}^t)\bigr) \\
&= \min\!\bigl(B,\, \min(k_{\mathrm{expand}}B,\, k_{\mathrm{expand}}^{t+1})\bigr).
\end{split}
\end{equation}

Because $k_{\mathrm{expand}}B \ge B$ for all integers $k_{\mathrm{expand}} \ge 1$, the inner minimum simplifies to $k_{\mathrm{expand}}^{t+1}$ whenever $k_{\mathrm{expand}}^{t+1} < B$, and otherwise the outer minimum enforces $B$. Thus
\[
|N_{t+1}| = \min(B, k_{\mathrm{expand}}^{t+1}).
\]
\qed 

Hence the frontier grows geometrically as $k_{\mathrm{expand}}^t$ until saturating at beam width $B$. An edge case is $k_{\mathrm{expand}}=1$, which yields a single chain $|N_t|=1$ (if $B \geq 1$).

In the presence of terminals or variable branching, the relation holds as an upper bound rather than an equality, i.e., $|N_t|\le \min(B,k_{\mathrm{expand}}^t)$.

\proposition{prop:cum-expansion-cost}
(Beam-search cumulative expansion cost.)
Assume each node generates at most $k_{\mathrm{expand}}$ children, pruning retains the top $B$ nodes after each layer, and no terminal states are encountered. Let $D\in\mathbb{N}$ be the maximum depth (number of layers). Define the per-depth expansion cost as the number of child-generation operations,
\begin{equation}
\begin{split}
C_{\mathrm{bs}}(t) 
&:= k_{\mathrm{expand}}\,|N_t| \\
&= k_{\mathrm{expand}}\,\min(B,\, k_{\mathrm{expand}}^t), \\
&\quad t=0,1,\ldots,D-1.
\end{split}
\end{equation}

Then the total expansion cost up to depth $D$ is

\begin{equation}
\begin{split}
C_{\mathrm{bs}}(D) 
&= \sum_{t=0}^{D-1} C_t \\
&= \sum_{t=0}^{D-1} k_{\mathrm{expand}}\,\min(B,\, k_{\mathrm{expand}}^t).
\end{split}
\end{equation}


\proposition{prop:cont-expansion-cost}
(Beam-search with chaining: cumulative expansion cost.)
Suppose chaining is enabled for the first $D_{C1}$ ``easy'' depths, where each expansion produces at most $k_{\mathrm{bn}}\le k_{\mathrm{expand}}$ children ($k_{\mathrm{bn}}=1$ for direct prompting, $k_{\mathrm{bn}}^t=k_{\mathrm{expand}}$ for the self-consistency approach). After depth $D_{C1}$, standard beam search resumes with branching factor $k_{\mathrm{expand}}$ and beam size $B$. Then the per-depth expansion cost is

Hence the total expansion cost up to depth $D$ is

\begin{equation}
\begin{aligned}
& C_{\mathrm{bs+chain}}(D) \\
&= \sum_{t=0}^{D_{C1}-1} k_{\mathrm{bn}} \\
&\quad + \sum_{t=D_{C1}}^{D-1}
    k_{\mathrm{expand}} \cdot
    \min\!\left(B,\, k_{\mathrm{expand}}^{\,t-D_{C1}}\right).
\end{aligned}
\end{equation}
\qed
\medskip

\subsection{Theoretical Error Analysis of BN Evaluation}
\label{app:bn_error_analysis}

\red{
\paragraph{BN only affects search through the \emph{chaining decision}.}
At a node with state $s_t$, BN evaluation produces a score $r_{\mathrm{bn}}$ and chaining is applied when $r_{\mathrm{bn}} \ge R_{\mathrm{bn}}$ (and, when applicable, $r_{\mathrm{conf}} < R_{\mathrm{conf}}$).
Therefore, BN errors can only change the search procedure by changing \emph{how often} nodes are classified as \emph{necessary}, i.e., how frequently chaining is used instead of standard expansion.
}

\red{
\paragraph{ToT-BS: BN errors shift the realized $D_{C1}$.}
Recall that $D_{C1}$ is defined as the first depth at which normal beam expansion resumes.
If BN is optimistic (branching is suppressed), more depths are classified as necessary and chaining persists longer, which increases the realized $D_{C1}$.
After depth $D_{C1}$, the frontier size follows the same form as in Appendix~\ref{app:theoretical_costs_bs}:
\[
|N_t| = \min\!\left(B,\, k_{\mathrm{expand}}^{\,t-D_{C1}}\right), \qquad t\ge D_{C1},
\]
so a larger $D_{C1}$ reduces the amount of branching available at a fixed depth $t$ and makes the search trajectory more chain-like.
If BN is pessimistic (branching is over-triggered), chaining stops earlier (smaller $D_{C1}$) and the procedure approaches the original ToT-BS.
}

\red{
Crucially, regardless of BN bias, the policy-invocation cost never increases:
as shown in Section~\ref{sec:theoretical_bs},
\[
C_{\mathrm{bs+chain}}(D) \;\le\; C_{\mathrm{bs}}(D),
\]
with strict reduction whenever some necessary depths use $k_{\mathrm{bn}}<k_{\mathrm{expand}}$.
Thus, overly optimistic BN may reduce exploration breadth (and potentially accuracy), but it does not harm efficiency relative to the baseline.
}

\red{
\paragraph{MCTS: BN errors change which first-expanded nodes are cheap.}
Under the \emph{full expansion on first visit} rule (Section~\ref{sec:theoretical_mcts}), the set $E(N)$ denotes the distinct nodes that are first-expanded within $N$ MCTS iterations, and $E^c(N)\subseteq E(N)$ denotes those among them classified as necessary by BN.
BN affects the expansion cost only through whether a first-expanded node uses $k_{\mathrm{bn}}$ (necessary) or $k_{\mathrm{expand}}$ (unnecessary), as shown in Equations \ref{eq:c_mcts} and \ref{eq:c_mcts_chain}.
Optimistic BN increases $|E^c(N)|$ (more first-expanded nodes are treated as necessary), which pushes the explored tree toward cheaper, narrower expansions.
Pessimistic BN decreases $|E^c(N)|$, which reduces chaining opportunities and makes the procedure closer to baseline MCTS.
}

\red{
Note that $|E(N)| \le N$ always holds, and $|E(N)|$ can be strictly smaller than $N$ when iterations frequently terminate early (e.g., reaching terminal states) or hit the depth limit $D$ before encountering a new unexpanded node.
This phenomenon is independent of BN and does not affect the monotonic guarantee below.
}

\red{
\paragraph{Guarantee under BN error.}
For both ToT-BS and MCTS, the efficiency guarantees hold regardless of BN bias because the design enforces $k_{\mathrm{bn}} \le k_{\mathrm{expand}}$.
In particular, for MCTS we always have
\[
C_{\text{mcts+chain}}(N) \;\le\; C_{\text{mcts}}(N),
\]
with strict inequality whenever at least one ```necessary'' node is encountered (i.e., $|E^c(N)|>0$).
Therefore, BN errors primarily trade off \emph{exploration breadth} versus \emph{cost savings}, while never increasing the number of policy invocations relative to the baseline.
}

\red{
\paragraph{Takeaway.}
Using only existing quantities, BN evaluation errors map to downstream search behavior through $(i)$ the realized $D_{C1}$ in ToT-BS and $(ii)$ the realized subset $E^c(N)$ of first-expanded nodes in MCTS.
A full theory that further maps these structural changes to \emph{solution accuracy} would require modeling how reduced branching affects the probability of encountering a correct trajectory within depth limit $D$ (or within $N$ iterations), which we leave as future work and instead study empirically in Section~\ref{sec:analysis} and Appendix~\ref{app:blocksworld}.
}

\section{Details of RAP and REST-MCTS}
\label{app:alg_details}

\subsection{Task Formulation: RAP vs ReST-MCTS}
\label{app:task_form_rap_vs_rest}
Table~\ref{tab:concate_vs_qa} shows the distinction between task formulation.
\begin{table*}[ht!]
    \centering
    \begin{tabular}{p{7cm}p{7cm}}
    \toprule
    Reasoning via Concatenation from Rest-MCTS 
    &  Reasoning via QA from RAP 
    \\\midrule
    First, calculate the total number of eggs used by Janet each day for breakfast and baking.\textcolor{red}{(Homogeneous Thought 1)} \newline
    Janet uses 3 eggs for breakfast and 4 eggs for baking, so the total number of eggs used each day is $3 + 4 = 7$.  \textcolor{red}{(Homogeneous Thought 2)}  
    &  What is the total number of eggs used by Janet each day for breakfast and baking? \textcolor{red}{(Question as action)} \newline
    Janet uses 3 eggs for breakfast and 4 eggs for baking, so the total number of eggs used each day is $3 + 4 = 7$.  \textcolor{green}{(Answer to form the next state)} 
    \\ \bottomrule
    \end{tabular}
    \caption{Formatting Difference Between Reasoning via Concatenation and QA under the same meaning.}
    \label{tab:concate_vs_qa}
\end{table*}

\subsection{Four Phases}
As standard MCTS, four phases are included for RAP and REST-MCTS.
\begin{itemize}
    \item Selection: 
    Walk down the tree until a leaf or an unexpanded node is reached.
    UCT values will be used for selection when all the child nodes have been visited (i.e., node.state is not empty).
    Otherwise , the reward is estimated to select from unvisited nodes.

    \item Expansion:
    If the leaf is not terminal and not depth-limited, the selected node will be expaned by calling the policy. All its $k$ children will be expanded at once.
    \begin{itemize}
        \item Lazy Sampling: Following RAP implementation \citep{hao-etal-2023-reasoning}, lazy sampling \citep{guez2012lazysampling} is used. After the policy generates multiple actions, RAP may leave action nodes unvisited for efficiency and only use the transition model to infer the next state if it is selected for expansion. Therefore, the sampled candidate nodes only maintain actions.

        \item Non-empty state may reflect the existence of value: Once the state of a node is not empty, the node must have gone through backpropagation and thus contain rewards for selection. But this does not be the case for continuous nodes due to the requirement of inferring the state during continuation.
    \end{itemize}

    \item Simulation (rollout):
    At each step, $k$ actions are sampled from policy, and then most reward models are used to select the most valuable one. 
    \begin{itemize}
        \item In RAP, the sampling process will proceed iteratively until a terminal step is reached or a global depth limit is reached.
        
        \item In ReST-MCTS, a fixed depth limit 2 is set specifically for roll-out. 
    \end{itemize}
    
    \item MCTS Backpropagation - Value Update:
    After each simulation returns a reward $r$, update the Q value as:
    \begin{equation}
        Q_{\text {new }}=\frac{r+Q_{\text {old }}  \cdot \text{Count}_{\text {new }}}{\text{Count}_{\text {new }}}, 
    \end{equation}
     $r$, depending on the task, can be a reward at the terminal state. In some cases,it can be an aggregated one, if each simulation step yields a reward. Specifically, if rewards \(r_t\) are discounted by \(\gamma\), then the final sample reward \(r\) for backpropagation is:
        \begin{equation}
        r = G = \sum_{t=0}^{T-1} \gamma^t r_t.
        \end{equation}
    These rewards from simulated nodes are then employed to update the Q values for non-simulated nodes, including the leaf nodes and those above.
    \( Q_{\text{old}} \) are the previous \( Q \)-value, and \( \text{Count}_{\text{new}} \) is the total visit count after the current update.

    During our implementation, each reward $r \in R_\text{cum}$ propagating from the terminal node is stored in a list $R_\text{cum}$ , and average them when used.
    The procedure of value estimate, provides the initialized values for new actions or resulting states.  

    For RAP, the confidence of transition models is also included along with the usefulness score to be processed for backpropagation
    \begin{equation}
       r^w \cdot r_{\text{conf}}^{(1 - w)}, 
    \end{equation}
    where $ w $ is the weight between 0 and 1.

    \item MCTS Backpropagation - Visit Update:
    Except for the estimated value $Q(s, a)$, the backpropagation also updates the visit count for every state on the path from the root to the leaf and each edge $(s, a)$ along that path, denoted as $\text{Count}\left(s_t\right)$ and $\text{Count}\left(s_t, a_{t+1}\right)$, respectively.
\end{itemize}


\section{Reward Models}
\label{app:lmpr}
We profile LLMs as reward models, as RAP does. However, the original profiling only consider the contribution of solution steps to the input task. 
As suggested by \citet{zhang2024restmcts}, we further improve the prompt to reflect the probability of correctness.
Three constraints are enforced: bounded range, contribution-awareness, and correctness-awareness in the original paper.  

\section{Incompatibility of Chat-Formatted LLMs with RAP}
\label{app:no_chat_in_rap}
Firstly, the policy and transition model in RAP requires a raw completion interface: the LLM is repeatedly invoked with concatenated sub-questions and answers, such as
\texttt{``Subquestion 1: ... Answer 1: ... Subquestion 2: ...''}.
This mismatch often leads to unstable generations (e.g., empty outputs, spurious punctuation, or off-task continuations), since the evolving transcript no longer resembles the conversational templates seen in training. 
Furthermore, the reward model requires the next token logically requiring \texttt{``yes''} or ~\texttt{``no''} for judgments (e.g., \texttt{``Is the new question useful?''}), however, the next token in the chat models tends to be a template one, which disrupts the intended binary evaluation.


\section{Parameters for LLM Inference}
\label{app:llm_params}

\paragraph{GPU Specification}

All experiments with LLaMA3 were conducted on NVIDIA A100-SXM4 40GB GPUs, while Qwen3 experiments were run on NVIDIA L40S 48GB GPUs.

\paragraph{Maximum Length.}
The maximum context length is set to 32,768 tokens (the upper limit of Qwen3) for all LITS roles and for CoT.  
In practice, 2,048 tokens are sufficient for all GSM8K instances, whereas Math500 often requires longer inputs and extended reasoning chains.  
For the policy models in ReST and ToT-BS, we instruct the model to keep each reasoning step within 1,000 tokens.  

For BN evaluation, we enforce a maximum of 1,000 output tokens for the aggregator model used in BN-SC$^1$.  
For $\text{LLM}_{\mathrm{eq}}$ (BN-SC$^2$) and $\text{LLM}_{\mathrm{bn}}$ (BN-DP), the same limit is applied, but only single-word outputs are accepted: “yes/no” for $\text{LLM}_{\mathrm{eq}}$, or a number between 1–4 for $\text{LLM}_{\mathrm{bn}}$. Special tokens are ignored in this check.  

\paragraph{Temperature.}
We follow the settings from prior work: RAP uses temperature 0.8 for LITS roles, ReST uses 0.7, and ToT-BS reuses the ReST policy with temperature 0.7.


    

\section{Raw Efficiency and Effectiveness Results}
\label{app:raw_tables}
For completeness and reproducibility, we report the raw numbers underlying the relative efficiency tables in the main paper. These include input tokens, output tokens, number of invocations, policy runtime, total runtime, and accuracy.

\begin{table*}[!ht]
\centering

\begin{subtable}{\textwidth}
\centering

\begin{tabular}{
    >{\raggedright\arraybackslash}p{2.2cm}
    >{\raggedleft\arraybackslash}p{0.85cm} 
    >{\raggedleft\arraybackslash}p{0.85cm} 
    >{\raggedleft\arraybackslash}p{0.85cm} 
    >{\raggedleft\arraybackslash}p{1cm} 
    >{\raggedleft\arraybackslash}p{0.85cm} 
    >{\raggedleft\arraybackslash}p{0.85cm} 
    >{\raggedleft\arraybackslash}p{0.85cm} 
    >{\raggedleft\arraybackslash}p{0.9cm} 
    >{\raggedleft\arraybackslash}p{0.9cm} 
    >{\raggedleft\arraybackslash}p{0.85cm} 
}
\toprule
 & \multicolumn{5}{c}{GSM8K} & \multicolumn{5}{c}{Math500} \\
\cmidrule(r){2-6} \cmidrule(l){7-11}
Method & Out & Inv & Time & Total & Acc & Out & Inv & Time & Total & Acc \\
\midrule

\textbf{ToT-BS} 
& 75.6k & 738 & 1.63H & 2.25H & 0.98 
& 594.5k & 1266 & 19.6H & 20.96H & 0.87 \\
\quad \textbf{+BN-SC$^1$} 
& 65.7k\rdn & 651\rdn & 1.39H\rdn & 1.89H\rdn & 0.96\rdn 
& 423.3k\rdn & 1230\rdn & 12.40H\rdn & 14.14H\rdn & 0.89\gup \\
\quad \textbf{+BN-SC$^2$} 
& 49.0k\rdn & 465\rdn & 1.01H\rdn & 1.25H\rdn & 0.96\rdn 
& 221.3k\rdn & 756\rdn & 5.34H\rdn & 6.03H\rdn & 0.84\rdn \\
\quad \textbf{+BN-DP} 
& 16.4k\rdn & 160\rdn & 0.35H\rdn & 0.51H\rdn & 0.97\rdn 
& 125.5k\rdn & 253\rdn & 4.29H\rdn & 4.57H\rdn & 0.86\rdn 
\\ \addlinespace\hline\addlinespace

\textbf{ReST} 
& 83.5k & 836 & 1.98H & 1.98H & 0.97\rdn 
& 491.0k & 1574 & 11.98H & 11.99H & 0.87= \\
\quad \textbf{+BN-SC$^1$} 
& 61.2k\rdn & 605\rdn & 1.30H\rdn & 2.22H\gup & 0.97\rdn 
& 287.9k\rdn & 886\rdn & 7.55H\rdn & 8.97H\rdn & 0.84\rdn \\
\quad \textbf{+BN-SC$^{2}$}
& 48.7k\rdn & 468\rdn & 1.01H\rdn & 1.66H\rdn & 0.97\rdn
& 192.9k\rdn & 661\rdn & 4.93H\rdn & 6.15H\rdn & 0.85\rdn \\
\quad \textbf{+BN-DP} 
& 17.0k\rdn & 166\rdn & 0.36H\rdn & 0.39H\rdn & 0.97\rdn 
& 86.4k\rdn & 238\rdn & 2.06H\rdn & 2.11H\rdn & 0.88\gup 
\\ \addlinespace\hline\addlinespace

\textbf{CoT} 
& 46.2k & 100 & 0.96H & 0.96H & 0.96 
& 78.1k & 100 & 1.61H & 1.61H & 0.79 \\
\bottomrule
\end{tabular}
\caption{Qwen3 32B.}
\label{tab:effi_effe_qwen}

\end{subtable}

\vspace{1em}

\begin{subtable}{\textwidth}
\centering

\begin{tabular}{
    >{\raggedright\arraybackslash}p{2.2cm}
    >{\raggedleft\arraybackslash}p{0.85cm} 
    >{\raggedleft\arraybackslash}p{0.85cm} 
    >{\raggedleft\arraybackslash}p{0.85cm} 
    >{\raggedleft\arraybackslash}p{1cm} 
    >{\raggedleft\arraybackslash}p{0.85cm} 
    >{\raggedleft\arraybackslash}p{0.85cm} 
    >{\raggedleft\arraybackslash}p{0.85cm} 
    >{\raggedleft\arraybackslash}p{0.9cm} 
    >{\raggedleft\arraybackslash}p{0.9cm} 
    >{\raggedleft\arraybackslash}p{0.85cm} 
}
\toprule
 & \multicolumn{5}{c}{GSM8K} & \multicolumn{5}{c}{Math500} \\
\cmidrule(r){2-6} \cmidrule(l){7-11}
Method & Out & Inv & Time & Total & Acc & Out & Inv & Time & Total & Acc \\
\midrule
\textbf{ToT-BS} 
& 108.6k & 1509 & 0.89H & 6.91H & 0.79 
& 458.1k & 2816 & 3.76H & 14.68H & 0.39 
\\

\quad \textbf{+BN-SC$^1$} 
& 124.8k\gup & 1771\gup & 1.01H\gup & 5.42H\rdn & 0.71\rdn 
& 617.0k\gup & 3877\gup & 5.14H\gup & 18.42H\gup & 0.35\rdn 
\\

\quad \textbf{+BN-SC$^2$} 
& 84.3k\rdn & 1171\rdn & 0.71H\gup & 1.32H\rdn & 0.64\rdn 
& 281.2k\rdn & 2119\rdn & 2.31H\rdn & 3.44H\rdn & 0.30\rdn \\

\quad \textbf{+BN-DP} 
& 33.9k\rdn & 438\rdn & 0.28H\rdn & 1.84H\rdn & 0.73\rdn 
& 137.8k\rdn & 1032\rdn & 1.14H\rdn & 4.58H\rdn & 0.27\rdn 
\\

\quad \textbf{+BN-SC$^1$+} 
& 105.4k\rdn & 1533\gup & 0.88H\rdn & 7.07H\gup & 0.80\gup 
& 627.1k\gup & 3534\gup & 5.28H\gup & 23.92H\gup & 0.38\rdn \\

\quad \textbf{+BN-SC$^2$+} 
& 76.9k\rdn & 1058\rdn & 0.64H\rdn & 1.88H\rdn & 0.76\rdn 
& 830.0k\gup & 3523\gup & 7.05H\gup & 13.96H\rdn & 0.39(=) 
\\
\quad \textbf{+BN-DP+} 
& 39.1k\rdn & 460\rdn & 0.33H\rdn & 2.01H\rdn & 0.77\rdn 
& 284.9k\rdn & 1751\rdn & 2.37H\rdn & 8.83H\rdn & 0.36\rdn 
\\ \addlinespace \hline \addlinespace

\textbf{ReST} 
& 130.4k & 1828 & 1.06H & 10.44H & 0.80 
& 640.8k & 3976 & 5.41H & 25.98H & 0.37 
\\
\quad \textbf{+BN-SC$^1$-}  
& 100.7k\rdn & 1352\rdn & 0.83H\rdn & 6.74H\rdn & 0.70\rdn 
& 491.4k\rdn & 3281\rdn & 4.20H\rdn & 20.89H\rdn & 0.33\rdn 
\\

\quad \textbf{+BN-SC$^{2}$-} 
& 68.0k\rdn & 960\rdn & 0.56H\rdn & 3.21H\rdn & 0.73\rdn
& 238.5k\rdn & 1944\rdn & 1.98H\rdn & 7.53H\rdn & 0.29\rdn 
\\

\quad \textbf{+BN-DP-} 
& 29.4k\rdn & 406\rdn & 0.25H\rdn & 2.63H\rdn & 0.68\rdn 
& 112.8k\rdn & 731\rdn & 0.95H\rdn & 5.04H\rdn & 0.36\rdn \\

\quad \textbf{+BN-SC$^1$+} 
& 76.9k\rdn & 1082\rdn & 0.63H\rdn & 6.67H\rdn & 0.84\gup 
& 221.1k\rdn & 1722\rdn & 1.81H\rdn & 12.87H\rdn & 0.43\gup 
\\

\quad \textbf{+BN-SC$^{2}$+} 
& 65.0k\rdn & 875\rdn & 0.54H\rdn & 3.38H\rdn & 0.80=
& 386.0k\rdn & 2053\rdn & 3.20H\rdn & 12.01H\rdn & 0.34\rdn 
\\
\quad \textbf{+BN-DP+}
& 27.6k\rdn & 354\rdn & 0.23H\rdn & 2.47H\rdn & 0.76\rdn 
& 213.4k\rdn & 1285\rdn & 1.78H\rdn & 9.10H\rdn & 0.37$=$ 
\\ \hline \addlinespace
\textbf{CoT} 
& 20.0k & 100 & 0.16H & -- & 0.68
& 121.5k & 316 & 1.01H & -- & 0.34 \\
\bottomrule
\end{tabular}
\caption{LlaMa3 8B Instruction. \textbf{BN-SC-}, \textbf{BN-DP-} uses the incompetent small LLaMA3-8B for the action aggregator or BN judge, respectively, whereas \textbf{BN-SC+} and \textbf{BN-DP+} employ the stronger Qwen3-32B for these roles.}
\label{tab:effi_effe_llama}
\end{subtable}
\label{tab:effi_effe}
\caption{Efficiency and effectiveness of of ToT-BS (or ReST-MCTS*) +CiT on GSM8K and Math500.  
\textbf{Out} = output tokens (policy),  
\textbf{Inv} = number of model invocations (policy),  
\textbf{Time} = wall-clock running time in hours (policy),  
\textbf{Time (Total)} = overall LLM running time in hours (LLMs as policy, reward models, transition models and BN evaluators),  
\textbf{Acc} = accuracy.
}
\end{table*}

\begin{table*}[!ht]
\centering

\begin{tabular}{
    >{\raggedright\arraybackslash}p{2.2cm}
    >{\raggedleft\arraybackslash}p{0.85cm} 
    >{\raggedleft\arraybackslash}p{0.85cm} 
    >{\raggedleft\arraybackslash}p{0.85cm} 
    >{\raggedleft\arraybackslash}p{1cm} 
    >{\raggedleft\arraybackslash}p{0.85cm} 
    >{\raggedleft\arraybackslash}p{0.85cm} 
    >{\raggedleft\arraybackslash}p{0.85cm} 
    >{\raggedleft\arraybackslash}p{0.9cm} 
    >{\raggedleft\arraybackslash}p{0.9cm} 
    >{\raggedleft\arraybackslash}p{0.85cm} 
}
\toprule
 & \multicolumn{5}{c}{GSM8K} & \multicolumn{5}{c}{Math500} \\
\cmidrule(r){2-6} \cmidrule(l){7-11}
Method & Out & Inv & Time & Total & Acc & Out & Inv & Time & Total & Acc \\
\midrule
\textbf{RAP} 
& 67.8k & 4606 & 0.69H & 5.53H & 0.61 
& 70.4k & 3583 & 0.67H & 8.03H & 0.18 
\\
\quad \textbf{+BN-SC+} 
& 23.5k\rdn & 1388\rdn & 0.24H\rdn & 2.88H\rdn & 0.48\rdn 
& 56.7k\rdn & 2347\rdn & 0.61H\rdn & 7.84H\rdn & 0.27\gup \\
\quad \textbf{+BN-SC$^{2}$+} 
& 14.9k\rdn & 1128\rdn & 0.15H\rdn & 1.06H\rdn & 0.46\rdn 
& 25.8k\rdn & 1449\rdn & 0.26H\rdn & 3.18H\rdn & 0.21\gup \\
\quad \textbf{+BN-DP+} 
& 7.7k\rdn & 625\rdn & 0.08H\rdn & 1.25H\rdn & 0.57\rdn 
& 14.7k\rdn & 697\rdn & 0.14H\rdn & 3.15H\rdn & 0.26\gup 
\\ \addlinespace \hline \addlinespace
\textbf{CoT} 
& 69.4k & 100 & 0.57H & -- & 0.32 
& 71.2k & 100 & 0.57H & -- & 0.17 \\
\bottomrule
\end{tabular}
\caption{Efficiency and effectiveness of RAP+CiT on GSM8K and Math500.}
\label{tab:effi_effe_rap}
\end{table*}

\section{Additional Failure Analyses}
\label{app:failure}

\subsection{Instance-Level Visualization}
In this section we provide the full set of instance-level visualizations for all failure cases of BN methods. 
Each figure reports the difference in the number of invocations between BN methods and the baseline across 100 instances, 
with filled markers indicating correct predictions and hollow markers incorrect ones.

\begin{figure*}[t]
    \centering
    \includegraphics[width=0.89\linewidth]{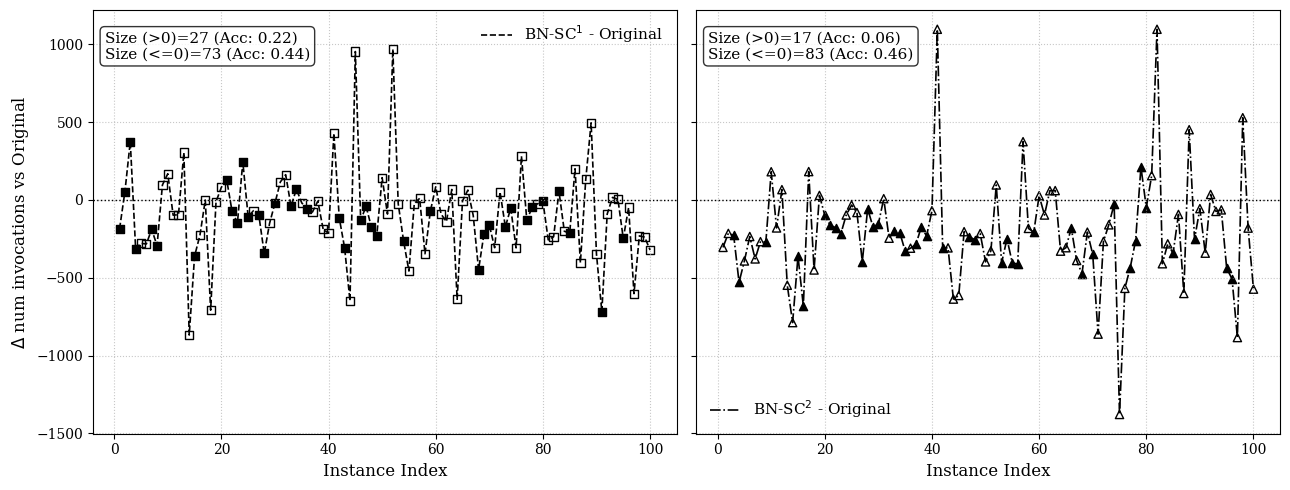}
    \caption{Instance-level analysis of a failure case on Math500 with LLaMA+Qwen (policy = LLaMA, BN = Qwen). 
Each point shows the relative change in the number of invocations compared to the baseline; negative values indicate higher efficiency (fewer output tokens required).  Filled markers correspond to correct predictions, while hollow markers correspond to incorrect ones.}
\label{fig:fail_math500_llama_qwen}
\end{figure*}

\begin{figure*}[!ht]
    \centering
\includegraphics[width=0.6\linewidth]{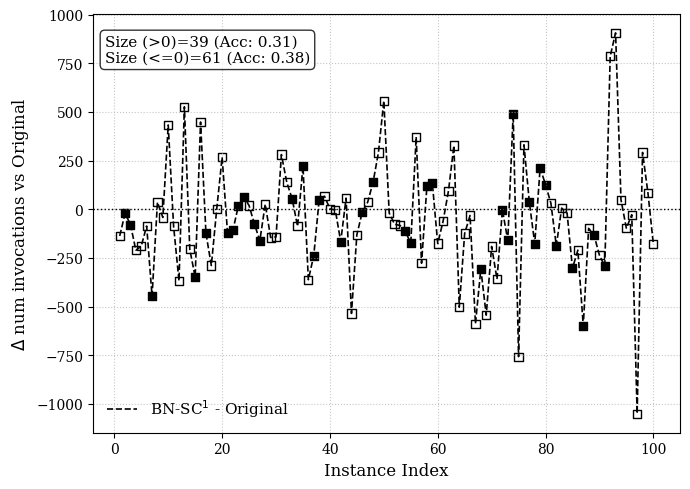}
    \caption{Instance-level analysis of failure for Math500 with LLaMA+LLaMA (BN-SC$^1$).}
    \label{fig:fail_math500_llama_llama}
\end{figure*}

\begin{figure*}[!ht]
    \centering
    \includegraphics[width=0.6\linewidth]{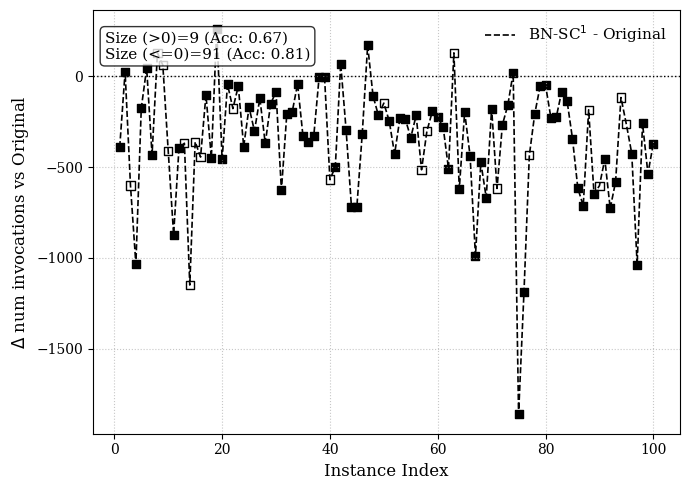}
    
    \caption{Instance-level analysis of failure for GSM8K with LLaMA+Qwen (BN-SC$^1$).}
    \label{fig:fail_gsm8k_llama_qwen}
\end{figure*}

\begin{figure*}[!ht]
    \centering
    \includegraphics[width=0.6\linewidth]{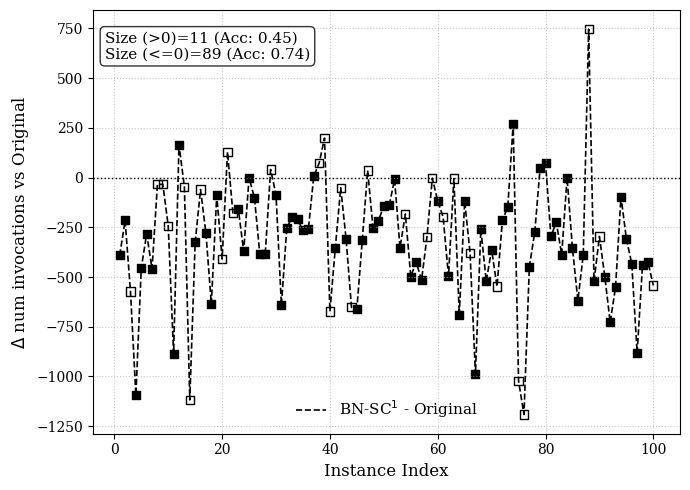}
    \caption{Instance-level analysis of failure for GSM8K with LLaMA+LLaMA (BN-SC$^1$).}
    \label{fig:fail_gsm8k_llama_llama}
\end{figure*}



\subsection{CiT Regression}
\label{app:cit_regression}
\red{
We analyze the correlation between reasoning tree structure and policy token overhead across 
different Chain-in-Tree (CiT) configurations. Token overhead is defined as the difference 
in policy output tokens between CiT and baseline BFS (CiT minus BFS), where positive values 
indicate efficiency regressions. Note that token overhead measures only policy invocations; 
the cost of BN evaluators is not included in this analysis.
}

\red{
We examine three metrics as potential predictors of token overhead:
\begin{itemize}
    \item \textbf{max\_depth}: The maximum depth of the reasoning tree, representing the 
    longest chain of reasoning steps explored.
    \item \textbf{branching\_factor}: The average number of nodes per depth level, 
    representing the width of exploration at each step.
    \item \textbf{level}: The problem difficulty rating (1--5 for MATH500, not applicable 
    for GSM8K), tested using Spearman rank correlation.
\end{itemize}
}

\red{
Table~\ref{tab:cit_correlations} summarizes the correlation between these metrics and 
token overhead across four experimental configurations. We use BN-SC$^1$ to denote the 
aggregator-based self-consistency method and BN-SC$^2$ for the pairwise self-consistency 
method. The superscript $^-$ indicates using an unreliable BN evaluator (Llama 8B, same 
as the policy model), while $^+$ indicates a reliable evaluator (Qwen 32B).
}
\begin{table*}[h]
\centering

\begin{tabular}{lcccc}
\toprule
\textbf{Configuration} & \textbf{Dataset} & \textbf{max\_depth ($r$)} & \textbf{branching\_factor ($r$)} & \textbf{level ($\rho$)} \\
\midrule
BN-SC1$^-$ (Llama 8B BN) & GSM8K & 0.796*** & 0.681*** & 0.072 \\
BN-SC1$^+$ (Qwen 32B BN) & MATH500 & 0.294** & 0.357*** & 0.086 \\
BN-SC1$^-$ (Llama 8B BN) & MATH500 & 0.248* & 0.267** & 0.084 \\
BN-SC2$^+$ (Qwen 32B BN) & MATH500 & 0.436*** & 0.146 & 0.117 \\
\bottomrule
\end{tabular}
\caption{Correlation between tree structure metrics and token overhead across CiT configurations. * $p < 0.05$, ** $p < 0.01$, *** $p < 0.001$}
\label{tab:cit_correlations}
\end{table*}

\red{
The results reveal several consistent patterns across all configurations:
}

\red{
\paragraph{Tree structure predicts overhead, problem difficulty does not.} 
Across all four configurations, max\_depth and branching\_factor show significant positive 
correlations with token overhead ($p < 0.05$), while problem difficulty level shows no 
significant correlation ($\rho \approx 0.08$--$0.12$, $p > 0.24$ in all cases). This 
indicates that efficiency regressions are driven by search dynamics rather than inherent 
problem characteristics. A difficult problem does not necessarily lead to higher overhead 
if the search terminates efficiently; conversely, an easy problem can cause regression if 
the continuation mechanism fails to recognize early termination opportunities.
}

\red{
\paragraph{BN evaluator reliability has limited impact on correlation patterns.}
Comparing BN-SC$^1$$^+$ ($r = 0.294$) with BN-SC$^1$$^-$ ($r = 0.248$) on the same MATH500 
dataset, we observe similar correlation strengths regardless of evaluator reliability. 
This suggests that the relationship between tree structure and policy token overhead is 
relatively stable across evaluator configurations. The slightly higher correlation with 
the reliable evaluator may indicate that better continuation decisions lead to more 
consistent (and thus more predictable) search behavior, though the difference is modest.
}

\red{
\paragraph{Dataset characteristics influence correlation magnitude.}
BN-SC$^1$$^-$ on GSM8K exhibits substantially stronger correlations ($r = 0.796$ for 
max\_depth, $r = 0.681$ for branching\_factor) compared to all MATH500 configurations 
($r = 0.25$--$0.44$). This difference likely reflects GSM8K's more uniform problem 
structure: simpler arithmetic problems lead to more predictable search patterns, where 
tree size directly translates to token usage. MATH500's diverse problem types 
(algebra, geometry, number theory, etc.) introduce greater variance in how tree 
structure relates to computational cost.
}


\red{
\paragraph{Implications}
While tree structure metrics show promise as indicators of efficiency 
regression risk, their practical application for early termination requires 
problem-specific calibration and further investigation into the distributional 
characteristics of reasoning trajectories.
We note several 
important caveats:
\begin{itemize}
    \item The appropriate thresholds for early termination are likely to be 
    \textbf{problem-dependent} and \textbf{dataset-dependent}. The optimal depth limit 
    for GSM8K problems may differ substantially from that for MATH500 problems due to 
    differences in problem complexity and solution structure.
    \item Establishing concrete early-stopping criteria would require additional analysis 
    of the \textbf{distribution of step lengths} across successful and unsuccessful 
    problem instances, which is beyond the scope of the current analysis.
    \item The correlations reported here are \textbf{post-hoc observations} on completed 
    search trajectories. Whether these metrics can be effectively used for \textbf{online 
    early detection} during search remains an open question that requires further 
    empirical validation.
\end{itemize}
}

\section{LLM Usage}
\label{app:llm_usage}
Large Language Models (LLMs) were used in this work as \textbf{writing assistants} to support clarity and style. Their role was confined to the revision stage of the manuscript and did not extend to research ideation, algorithm design, proofs, experiments, or interpretation of results.  

The revision process followed a structured, iterative workflow:  

\begin{enumerate}
    \item \textbf{Abstract revision:}  
    \begin{itemize}
        \item The LLM was asked to revise the abstract.  
        \item The authors manually revised the generated text, compared it against the original version, and selectively incorporated improvements.  
        \item For specific sentences, the LLM was prompted in-quote to suggest localized adjustments (e.g., rephrasing for clarity or conciseness).  
        \item The finalized abstract was authored by the authors after reviewing and merging both versions.  
    \end{itemize}

    \item \textbf{Section-by-section revision:}  
    \begin{itemize}
        \item With the finalized abstract fixed, the same process was repeated sequentially for each section: Section~1, Section~2, Section~3, Section~4, Section~5, Section~6, and the Appendix.  
        \item At each stage: (i) the LLM was asked to revise the draft section, (ii) the authors manually revised and compared the LLM output with the original draft, and (iii) the LLM was used for fine-grained, sentence-level in-quote adjustments where needed.  
        \item The revised section was only finalized after careful author editing and approval.  
    \end{itemize}

    \item \textbf{Formatting assistance:}  
    \begin{itemize}
        \item The LLM was occasionally used to generate LaTeX table skeletons, figure captions, and consistent notation formatting. All outputs were verified and edited by the authors.  
    \end{itemize}
\end{enumerate}

This procedure ensured that the LLM functioned only as a \textbf{linguistic and formatting assistant}. All scientific content---including the conception of the \textit{Chain-in-Tree} framework, technical derivations, algorithmic innovations, and empirical analysis---originated entirely from the authors.  

The authors take \textbf{full responsibility} for the correctness, originality, and integrity of all content.


\section{Package Design}
\label{app:package_design}

The LangAgent Python package modularizes LLM-profiled roles across various search frameworks.


\paragraph{Role Modularity.}
Specifically,
LangAgent organizes all LLM-mediated reasoning roles behind stable interfaces so that different search frameworks can reuse the same components without behavioral drift. 
Core contracts for the \emph{world model}, \emph{policy}, and \emph{reward/evaluator} roles are declared once in \texttt{langagent/reasoner\_base.py}, together with the shared \texttt{BaseSearchConfig} structure that wires termination and continuation settings across algorithms.

Concretely, \texttt{WorldModel} methods are defined for initialization, state transitions, and terminal checks. 
The policies inherit from the common \texttt{QAPolicy} skeleton, while evaluator variants specialize \texttt{QAEvaluator.fast\_reward} to add correctness or usefulness signals. 

\paragraph{Functional Search Pipelines.}
Search algorithms in \texttt{langagent/search} follow a functional design that keeps each phase explicit and composable. 
Breadth-first search exposes pure helpers for expansion, continuation-aware expansion and terminal checks that the top-level \texttt{bfs\_topk} function orchestrates. 
Likewise, the Monte Carlo tree search implementation separates selection, expansion, simulation, and backpropagation.

Identical role objects can drive either algorithm, and new planners can reassemble the same primitives to instrument alternative search behaviors without duplicating logic.

\paragraph{Utilities.}
Auxiliary utilities—answer extraction, dataset loading, and evaluation—live in \texttt{langagent/langreason/common.py}, ensuring that any new framework automatically gains the same output handling.
\section{BlocksWorld Experiments}
\label{app:blocksworld}
\subsection{\red{Experimental Setting}}

\red{
\paragraph{Setup of Tree Search.}
RAP has been employed in the BlocksWorld evaluation, while ToT-BS is not included because it was not designed
for action-based planning tasks and has not been adapted to BlocksWorld-style
environments in prior work.}

\red{
RAP requires a scalar reward or value signal to guide tree search, but does not
assume a specific implementation.
While the original RAP \citep{hao-etal-2023-reasoning} derives this signal from token-level logits, such access
is unavailable in API-based LLM settings.
We therefore adopt an explicit LLM-based step evaluator that directly scores
state–action pairs.
This evaluator serves the same functional role as a reward model and is a
standard design choice in planning-oriented LLM evaluations \citep{zhou2024language,yao2023tree} where internal
model logits are inaccessible.}

\red{
To encourage exploration within MCTS, we make two modifications to the default RAP settings.
First, we set the self-consistency (SC) threshold to 0.99, so that the model
continues chaining only when the policy consistently generates the same action;
otherwise, branching is triggered.
Second, we increase the LLM temperature to 1.
All other hyperparameters are kept identical to those used in language
reasoning tasks.
Despite these changes, we observe that BN-SC still exhibits limited exploration
in planning-oriented tasks such as BlocksWorld.
}

\red{
\paragraph{Use of Claude Models}
For the BlocksWorld experiments, we use Claude models accessed via a standard
API interface due to resource availability.
Since CiT is model-agnostic and operates purely at the inference level,
the choice of backbone LLM does not affect the applicability of the method.
Accordingly, we do not compare absolute performance across different models,
and instead focus on \textbf{relative efficiency gains within the same backbone}.}

\red{
\paragraph{Testing Data}
Due to API cost constraints, we evaluate all methods on a random subset of 10
BlocksWorld instances.
While this sample size is insufficient for drawing strong statistical
conclusions about absolute performance, it is adequate for our purpose of
analyzing \emph{relative efficiency–accuracy trade-offs} under a fixed model
and environment.
Since all methods are evaluated on the same instances with identical API
configurations, the observed reductions in token usage and inference cost are
stable and indicative of the core behavior of CiT-style search control.
}

\subsection{Results}
\red{
Table~\ref{tab:raw_blocksworld} reports raw token usage, inference cost, and
accuracy on BlocksWorld, while
Table~\ref{tab:rel_change_blocksworld} summarizes relative changes with respect to
RAP.
}

\red{
\paragraph{Efficiency.}
BN-SC achieves substantial efficiency gains over RAP.
Compared to RAP, BN-SC reduces input tokens by 79.9\% and output tokens by 83.8\%,
leading to an 81.2\% reduction in total API inference cost
(from \$8.86 to \$1.67).
This cost closely approaches that of the pure chaining baseline
(CoT), indicating that BN-SC effectively collapses the search tree
into a near-linear reasoning trajectory when branching is suppressed.
}

\red{
In contrast, BN-DP provides more moderate efficiency improvements,
reducing input tokens by 19.5\%, output tokens by 25.0\%, and total cost by
21.4\%, while still maintaining non-trivial search behavior.
}

\red{
\paragraph{Accuracy.}
BN-DP maintains most of the planning capability of RAP, achieving 40\% accuracy
compared to 50\% for RAP.
This suggests that explicit step-level evaluation can still guide effective
search, albeit with some loss in solution quality.
}

\red{
BN-SC converges to the performance of the chaining baseline, achieving 20\%
accuracy, identical to CoT.
This behavior reflects a regime where branching is largely suppressed and the
search trajectory becomes effectively linear.
}



\red{
\paragraph{Takeaway.}
Although BlocksWorld differs substantially from mathematical reasoning, these
results reinforce the same conclusions observed in the main experiments.
First, CiT remains effective in reducing the computational cost of tree search
by suppressing unnecessary branching.
Second, the effectiveness of CiT strongly depends on the quality of BN
evaluation: in this planning setting, BN-SC produces unreliable BN decisions,
which suppress branching excessively and consequently reduces the effectiveness
of CiT.
Importantly, these observations are consistent with the findings reported in
the main text.
}

\begin{table}[ht!]
\centering
\footnotesize
\begin{tabular}{
    >{\raggedright\arraybackslash}p{2.2cm}
    >{\raggedleft\arraybackslash}p{0.95cm} 
    >{\raggedleft\arraybackslash}p{0.95cm} 
    >{\raggedleft\arraybackslash}p{1cm}    
    >{\raggedleft\arraybackslash}p{0.85cm} 
}
\toprule
 & \multicolumn{4}{c}{BlocksWorld} \\
\cmidrule(r){2-5}
Method & In & Out & Cost & Acc \\
\midrule

\textbf{ReAct}
& 151{,}709
& 6{,}092
& \$0.5465
& 0.20
\\

\textbf{RAP}
& 1{,}960{,}749
& 198{,}543
& \$8.8604
& 0.50
\\

\quad \textbf{+BN-DP}
& 1{,}578{,}131
& 148{,}834
& \$6.9669
& 0.40
\\

\quad \textbf{+BN-SC}
& 393{,}993
& 32{,}252
& \$1.6658
& 0.20
\\

\bottomrule
\end{tabular}
\caption{Raw input/output token usage, API inference cost, and accuracy on BlocksWorld
using Claude~3.5 Sonnet.
These values are used to compute the relative changes in Table~\ref{tab:rel_change_blocksworld}.}
\label{tab:raw_blocksworld}
\end{table}

\begin{table}[ht!]
\centering
\footnotesize
\begin{tabular}{
    >{\raggedright\arraybackslash}p{2.2cm}
    >{\raggedleft\arraybackslash}p{0.95cm} 
    >{\raggedleft\arraybackslash}p{0.95cm} 
    >{\raggedleft\arraybackslash}p{1cm}    
    >{\raggedleft\arraybackslash}p{0.85cm} 
}
\toprule
 & \multicolumn{4}{c}{BlocksWorld} \\
\cmidrule(r){2-5}
Method & In & Out & Cost & Acc \\
\midrule

\multicolumn{5}{l}{\textbf{Relative to RAP} \; (baseline  Acc: \textbf{0.50}; ReAct: 0.20)} \\

\quad \textbf{BN-DP}
& 19.5\%\rdn
& 25.0\%\rdn
& 21.4\%\rdn
& 20.0\%\rdn
\\

\quad \textbf{BN-SC}
& 79.9\%\rdn
& 83.8\%\rdn
& 81.2\%\rdn
& 60.0\%\rdn
\\

\bottomrule
\end{tabular}
\caption{Relative changes in input tokens, output tokens, inference cost, and accuracy
compared to RAP on BlocksWorld using Claude~3.5 Sonnet.
\rdn\ denotes decrease.}
\label{tab:rel_change_blocksworld}
\end{table}

\section{Full Math500 Evaluation (316 Instances)}
\label{app:full_math500}

To verify that the efficiency--accuracy trends reported in the main experiments (based on the first 100 instances) are not driven by subset variance, we additionally evaluate on all 316 filtered Math500 instances using Qwen-32B under both ReST-MCTS and ToT-BS settings.

Table~\ref{tab:full_math500} reports the results. Under ReST-MCTS, BN-DP reduces total runtime from 49.43H to 9.16H ($\approx$5.4$\times$ reduction) while accuracy changes from 0.86 to 0.88. Under ToT-BS, runtime decreases from 42.42H to 8.69H ($\approx$4.9$\times$ reduction) while accuracy changes from 0.896 to 0.87. These results confirm that the efficiency gains and accuracy preservation observed on the 100-instance subset generalize to the full filtered benchmark.

\begin{table}[ht]
\centering
\footnotesize
\begin{tabular}{lcc}
\toprule
\textbf{Method} & \textbf{Total Runtime} & \textbf{Accuracy} \\
\midrule
\multicolumn{3}{l}{\textit{ReST-MCTS (Qwen-32B)}} \\
\quad Baseline & 49.43H & 0.86 \\
\quad +BN-DP   & 9.16H  & 0.88 \\
\midrule
\multicolumn{3}{l}{\textit{ToT-BS (Qwen-32B)}} \\
\quad Baseline & 42.42H & 0.896 \\
\quad +BN-DP   & 8.69H  & 0.87 \\
\bottomrule
\end{tabular}
\caption{Efficiency and accuracy on all 316 filtered Math500 instances with Qwen-32B. BN-DP achieves 4.9--5.4$\times$ runtime reduction with negligible accuracy change, consistent with the 100-instance results in the main paper.}
\label{tab:full_math500}
\end{table}

\end{document}